\pdfoutput=1

\documentclass[11pt]{article}

\usepackage[preprint]{acl}

\usepackage{times}
\usepackage{latexsym}

\usepackage[T1]{fontenc}

\usepackage[utf8]{inputenc}

\usepackage{microtype}

\usepackage{inconsolata}

\usepackage{graphicx}

\usepackage{dirtytalk}
\usepackage{amssymb}
\usepackage{amsmath}
\usepackage{subcaption}
\usepackage{float}

\usepackage{multirow}

\title{Leveraging Semantic Triples for Private Document Generation\\with Local Differential Privacy Guarantees}

\author{Stephen Meisenbacher, Maulik Chevli,\and Florian Matthes \\
Technical University of Munich\\
School of Computation, Information and Technology \\
Department of Computer Science\\
Garching, Germany\\
\texttt{\{stephen.meisenbacher,maulikk.chevli,matthes\}@tum.de} \\
}

\begin{document}
\maketitle
\begin{abstract}
Many works at the intersection of Differential Privacy (DP) in Natural Language Processing aim to protect privacy by transforming texts under DP guarantees. This can be performed in a variety of ways, from word perturbations to full document rewriting, and most often under \textit{local} DP. Here, an input text must be made indistinguishable from any other potential text, within some bound governed by the privacy parameter $\varepsilon$. Such a guarantee is quite demanding, and recent works show that privatizing texts under local DP can only be done reasonably under very high $\varepsilon$ values. Addressing this challenge, we introduce \textsc{DP-ST}, which leverages semantic triples for neighborhood-aware private document generation under local DP guarantees. Through the evaluation of our method, we demonstrate the effectiveness of the \textit{divide-and-conquer} paradigm, particularly when limiting the DP notion (and privacy guarantees) to that of a \textit{privatization neighborhood}. When combined with LLM post-processing, our method allows for coherent text generation even at lower $\varepsilon$ values, while still balancing privacy and utility. These findings highlight the importance of coherence in achieving balanced privatization outputs at reasonable $\varepsilon$ levels.
\end{abstract}

\section{Introduction}
\begin{figure*}[ht]
    \centering
    \includegraphics[scale=0.375]{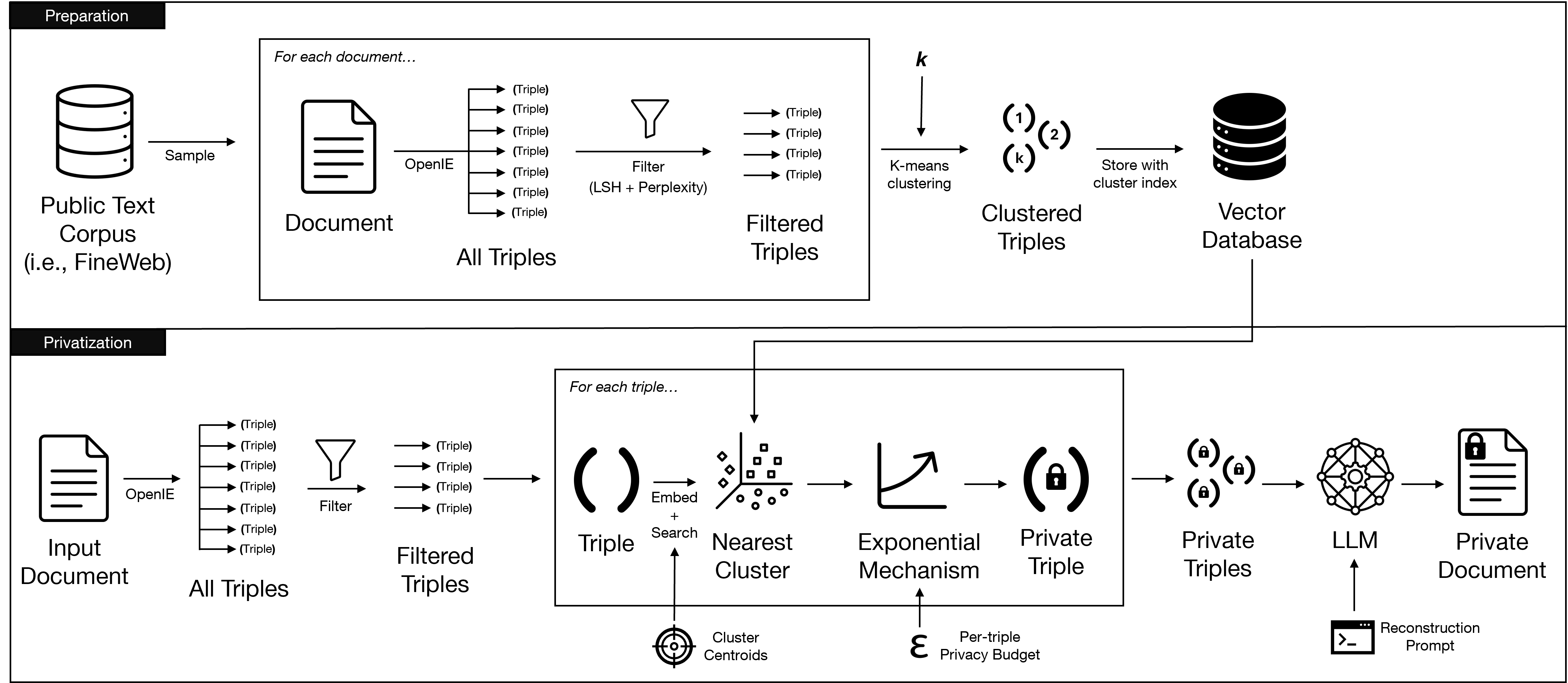}
    \caption{An overview of the \textsc{DP-ST} pipeline. In the \textit{Preparation} stage, publicly available texts are decomposed into semantic triples and stored in a vector database with cluster indices. Then, during \textit{Privatization}, each extracted triple from an input document is replaced with a differentially private triple achieved through the use of the Exponential Mechanism. All private triples are woven together into a reconstructed private document as the output.}
    \label{fig:DPST}
\end{figure*}

The field of \textit{text privatization} encompasses a variety of techniques \cite{sousa2023keep}, ranging from anonymization to differentially private text rewriting methods \cite{hu-etal-2024-differentially}. Such methods strive to balance the removal or perturbation of sensitive data and usefulness of the resulting text for the downstream applications, resulting in the so-called \textit{privacy-utility} tradeoff. Differential Privacy (DP) \cite{dwork2006differential}, being a mathematically grounded notion of privacy, provides a useful starting point for balancing the privacy-utility trade-off through the $\varepsilon$ parameter, or the \textit{privacy budget}.

The application of DP in text privatization can be performed at various lexical levels, such as words, sentences, or documents. Earlier works introduced the \textit{divide-and-conquer} paradigm, e.g., where the \textit{word} is chosen as the unit of protection, and full texts are privatized by composing word-level DP guarantees \cite{fernandes2019generalised,10.1145/3336191.3371856, carvalho2023tem, arnold-etal-2023-guiding}. Such composed texts, however, can lack coherence and grammatical structure \cite{mattern-etal-2022-limits}. 

Later works diverged from this paradigm and introduced methods for providing document-level guarantees, namely through DP noise addition on various text representations \cite{10.1145/3485447.3512232}, such as the latent vector in a \textsc{BART} model \cite{igamberdiev-habernal-2023-dp}. However, these approaches lack the control over which information to privatize, as opposed to word-level approaches. Moreover, they require a higher privacy budget to generate text that has reasonable utility since these works follow the strict local DP definition \cite{igamberdiev-habernal-2023-dp}. 

Recent works have leveraged the Exponential Mechanism (EM) \cite{4389483} in the generation step of language models, allowing for contextualized privatization and the ability to steer DP privatization with next token prediction probabilities \cite{bo-etal-2021-er, mattern-etal-2022-limits, utpala-etal-2023-locally, meisenbacher-etal-2024-dp}, thus providing a \textit{token}-level DP guarantee. However, these do not view the DP privatization task as a \textit{bounded} privacy budget problem, thereby exacerbating the issue of weak document-level privacy guarantees. Additionally, EM-approaches cannot guarantee grammatical correctness or coherence, due to randomization in the token prediction step.

We address these challenges by introducing \textsc{DP-ST}, a text privatization method with slightly relaxed local DP guarantees, which we call neighborhood-aware DP, for enhanced utility. \textsc{DP-ST} leverages a different \textit{divide-and-conquer} paradigm, by first decomposing input documents into only their \textit{core information}, represented by semantic subject-verb-object (SVO) triples \cite{schneider-etal-2024-comparative}, or simply, \textit{semantic triples}. Only these triples are privatized under local DP guarantees, aided by a corpus of clustered public semantic triples, whereby the privatization of a triple only considers the public triples within the most semantically similar cluster. Finally, \textsc{DP-ST} takes advantage of the post-processing principle of DP, in that privatized triples are woven together by an LLM to output a reconstructed private document.  

We find that using \textsc{DP-ST} for local DP text privatization allows for strong privacy protections \textit{and} utility preservation, while also doing so under low document-level privacy budgets. We show that in comparison to state-of-the-art methods, \textsc{DP-ST} performs considerably better in terms of ensuring coherent outputs and balancing the privacy-utility trade-off. Our experimental results demonstrate the impact of careful DP mechanism design, and we also highlight the importance of reasonable privacy budgets in local DP text privatization, providing a method that begins to reverse the trend of very large or even unbounded privacy budgets.

Our work contributes to the field of text privatization and DP text rewriting with the following:
\begin{enumerate}
    \itemsep -0.2em
    \item We propose \textsc{DP-ST}, which leverages semantic triple decomposition and LLM post-processing for coherent and privacy-preserving text generation under local DP.
    \item We demonstrate the effectiveness of \textsc{DP-ST} outputs in utility and privacy experiments, which outperforms SOTA methods in finding positive privacy-utility trade-offs.
    \item We open-source the code for triple corpus creation and clustering, as well as the \textsc{DP-ST} method. The repository is found at \url{https://github.com/sjmeis/DPST}.
\end{enumerate}

\section{Foundations and Related Work}
DP mechanisms lend plausible deniability to the input of the mechanism by adding calibrated noise to the output. Hence, an observed output cannot be attributed to a specific input with a high probability. Formally, for finite spaces $\mathcal{P}$ and $\mathcal{V}$ with $n$ and $m$ elements respectively, a randomized mechanism $\mathcal{M}: \mathcal{P} \to \mathcal{V}$ is a \textit{local} $\varepsilon$-DP mechanism iff $\forall x, y \in \mathcal{P}$ and $\forall z \in \mathcal{V}$, the following holds:
\[
    Pr[\mathcal{M}(x) = z] \leq e^{\varepsilon} Pr[\mathcal{M}(y) = z]
\]
The translation of DP into NLP (DP NLP) has increased in research interest in recent years, resulting in a great deal of works investigating primarily either model training with DP or DP text privatization \cite{hu-etal-2024-differentially}. The integration of DP into NLP, largely dealing with unstructured data, presents several initial challenges as the original notion of DP was designed for structured, tabular data \cite{klymenko-etal-2022-differential}. Beyond this, research in DP NLP carries an extra responsibility of showing that new methods can \textit{empirically} improve privacy protections, while also abiding by the \textit{theoretical} constraints of the DP definition.

The constraints of DP in the context of NLP have led to many challenging, yet interesting research directions. \citet{mattern-etal-2022-limits} highlight the difficulty in privatizing texts under DP constraints while also maintaining grammatical correctness and semantic coherence, while \citet{igamberdiev-habernal-2023-dp} echo this notion by demonstrating that such coherence can only be achieved at much higher privacy budgets. Other works point to crucial considerations that must be made in the design and evaluation of DP NLP methods, including ensuring adherence to DP \cite{habernal-2021-differential}, increasing transparency in evaluation \cite{igamberdiev-etal-2022-dp}, and considering the interdependence of language in datasets \cite{vu-etal-2024-granularity}.

\section{Method}
Our proposed method, named \textsc{DP-ST}, rests upon the idea that only the \textit{core} information of a sensitive input document should be privatized, both to focus on the key semantics as well as to reduce the number of DP perturbations needed. We realize this goal by relying on SVO triples, which we call \textit{semantic triples} for the remainder of this work. In the following, we outline our methodology for decomposing input texts into triples, privatizing these triples under local DP, and reconstructing the privatized triples to form coherent output documents.

\subsection{Decomposing a Text into Semantic Triples}
A \textit{semantic triple}, also known as an SVO or RDF triple, is a basic unit of the RDF framework\footnote{\url{https://www.w3.org/TR/PR-rdf-syntax/}} that expresses a relation between a \textit{subject} (\say{Mark Zuckerberg}) and an \textit{object} (\say{Facebook}), connected by some \textit{predicate} or \textit{verb} (\say{founded}). Together, the units within semantic triples form the basis for the key information communicated in any sentence.

\paragraph{Extracting Triples.}
As the first step of our pipeline, we construct an algorithm to extract semantic triples from a given document in a reliable and efficient manner. We leverage the Stanford Open Information Extraction (OpenIE) software, which comes as part of the Stanford CoreNLP Toolkit \cite{manning-etal-2014-stanford}. One particular feature allows for the open domain extraction of semantic triples \cite{angeli-etal-2015-leveraging}, which we use as the base extraction method. Concretely, we convert the extractions into a single string, for example:

\begin{center}
    \textit{Mark Zuckerberg} | \textit{founded} | \textit{Facebook}
\end{center}

Upon initial testing, we noticed that the extraction algorithm from OpenIE returns many redundant results, specifically overlapping spans which point to the same original SVO triple. Therefore, we implemented post-extraction filtering steps to narrow down the results to a set of distinct triples. We first utilize \textit{Locality-Sensitive Hashing} (LSH) \cite{10.5555/645925.671516} to quickly bucket similar triples. For this, we use the \textsc{MinHash} algorithm from the \textsc{datasketch} library \cite{eric_zhu_2024_11462182}, with a threshold of 0.4 and 128 permutations. Then, within each bucket, we perform perplexity filtering by choosing the triple with the lowest perplexity in each bucket. We do this to select only the most coherent triple, avoiding noisy extractions from OpenIE. We calculate perplexity using a GPT-2 model \cite{radford2019language}.

\paragraph{Building a Public Triple Corpus.}
With the triple extraction algorithm, we then proceeded to create a large-scale corpus of \say{public} semantic triples, garnered from open-source and publicly available text corpora. The purpose of this corpus was to serve as a public knowledge base of triples, which we can use later in our pipeline for DP privatization.

We use the \textsc{FineWeb} text corpus \cite{penedo2024the}, which consists of 15 trillion tokens of texts that were cleaned and deduplicated from Common Crawl data. In particular, we use the \textsc{sample-10BT} split, a smaller random subset of the larger corpus. We iterate through this subset, running our triple extraction algorithm with filtering, and storing the resulting triples into a Weaviate\footnote{\url{https://github.com/weaviate/weaviate}} vector database. As an embedding model, we employ \textsc{jina-embeddings-v3} \cite{sturua2024jinaembeddingsv3multilingualembeddingstask}, using 32-dimensional Matryoshka embeddings for search efficiency without the loss of quality.

We set a stopping criterion of 15 million extracted triples, which was reached after processing just under 800k texts from the \textsc{FineWeb} corpus. After deduplication, this resulted in roughly 13.4 million unique triples in our public triple database.

\paragraph{Clustering Triples.}
The final step in the preparation of our database of public triples involved a clustering process to group semantically related triples. This step was taken to align with our goal for \textit{neighborhood-aware} DP privatization, where an input triple is only to be privatized considering its most semantically related neighbors. This is introduced in more detail in Section \ref{sec:privatization}.

An important consideration was the \textit{number} of clusters to be formed, as we opted to use the efficient k-means clustering algorithm from \textsc{sklearn}. We made this $k$ value a privatization parameter, and we ran the clustering process for three $k$ values: 50k, 100k, and 200k. Intuitively, this would produce, on average, clusters with 268, 134, and 67 members, respectively. After each clustering process, we recorded for each vector in our database its three corresponding cluster IDs, marked as \textit{properties} in the database records. We stored all cluster centroids for use in the privatization step.

\subsection{DP Privatization of Semantic Triples}
\label{sec:privatization}
The first step in producing private documents is to decompose an input text into semantic triples using the same process as outlined above. Then, for each of these triples, a nearest-neighbor search is run by comparing the cosine similarity of the embedded triple string to each of the $k$ cluster centroids, depending on the chosen $k$ value. The nearest cluster, composed of all vectors belonging to the chosen centroid, is the \textit{privatization neighborhood}.

Within the privatization neighborhood, we model the application of DP as a \textit{selection} task, which lends itself to the use of the fundamental Exponential Mechanism (EM), as previously introduced. Specifically, the cosine similarity of the input triple in question is calculated with respect to each public triple in the neighborhood. The resulting scores are used in the EM, yielding a \textit{sensitivity} of 1 (i.e., the range of cosine similarity, see Appendix \ref{sec:proof-appendix}). The EM, which converts these scores into selection probabilities with a chosen $\varepsilon$ parameter, is run, yielding a private triple replacement.

After this process is followed for each extracted triple from the input document, the result is a set of output \textit{private triples}. It is crucial to note here that the DP guarantee lies on the \textit{triple} level and grants indistinguishability of input triples only within a neighborhood (cluster). Thus, in order to privatize an entire document under local DP, a privacy budget $\varepsilon$ must be distributed among all input triples. This is discussed further next, along with a proof that our method is differentially private.

\subsubsection{Clarifying the \textit{DP} in \textsc{DP-ST}}
\label{sec:proof}
For local DP, every element of a domain is a neighbor of every other element. For \textsc{DP-ST}, we relax this condition and consider only the triples belonging to a cluster as neighbors. Moreover, the codomain is also restricted to a single cluster. This significantly reduces the range of our mechanism to only the presumably semantically relevant triples. The classic Exponential Mechanism is used on the input triple to produce a privatized triple from a cluster of public triples. The full DP proof for this mechanism is deferred to Appendix \ref{sec:proof-appendix}. We note that our usage of the term \textit{local DP} with \textsc{DP-ST} is connected to the relaxed notion described here; this distinction is further discussed in Section \ref{sec:discuss}.

\subsection{Document Reconstruction via Post-processing with LLMs}
Unfortunately, a set of private triples does not lend itself well to downstream tasks or the sharing of coherent texts. As such, we devise a method for reconstructing documents from a collection of triples. Previous work \cite{schneider-etal-2024-comparative} has shown the effectiveness of LLMs in precisely this task, namely, text generation from semantic triples.

To realize document reconstruction, we leverage the \textit{post-processing} property of DP, which states that any arbitrary computations performed on DP outputs will still be DP. Formally, if $\mathcal{F}(x)$ is $\varepsilon$-DP, $g(\mathcal{F}(x))$ is also $\varepsilon$-DP, for any $g$. This becomes important for reducing the degrading effects of DP noise, and has been leveraged in various applications \cite{pmlr-v80-balle18a,census,10.1145/3664476.3669926}.

Under this premise, we utilize the generative capabilities of LLMs in order to fuse a set of private triples into one coherent text. We base our prompt on that used by \citet{schneider-etal-2024-comparative} in their experiments, and we make only slight modifications. The prompt can be found in Appendix \ref{sec:prompt}.

\section{Experimental Setup and Results}
We evaluate the efficacy of \textsc{DP-ST} in a series of experiments that test both its empirical privacy protections as well as its ability to preserve utility. All utilized datasets contain a privacy-sensitive task, namely, adversarial authorship or attribute inference. We divide these datasets into two groups: those with an associated downstream task and those without one. For the datasets without a \say{utility task}, we only apply automatic utility metrics. In both cases, the privacy and utility measurements allow for the quantification of the privacy-utility trade-off, forming the basis for a comparative analysis between \textsc{DP-ST} and selected SOTA methods.

\subsection{Datasets and Tasks}
\label{sec:datasets}
We use five datasets in total, ensuring diversity in dataset size, average document length, and domain.

\paragraph{Reuters News.} 
We use the \textit{Reuters\_50\_50} dataset \cite{reuter_50_50_217}, which is a subset of the much larger RCV1 corpus \cite{10.5555/1005332.1005345}. This subset contains 2500 news articles in total, with 50 articles from each of 50 authors. This creates a 50-class adversarial authorship identification task.

\paragraph{Spooky Authors.}
We use the dataset from the Spooky Author Identification challenge\footnote{\url{https://www.kaggle.com/c/spooky-author-identification/}}, which contains 19,579 excerpts from three authors (Edgar Allen Poe, HP Lovecraft, Mary Shelley). We transform this into an adversarial identification task.

\paragraph{Reddit Mental Health.}
We create a subset of a dataset containing Reddit posts covering the topic of mental health\footnote{\url{https://huggingface.co/datasets/solomonk/reddit_mental_health_posts}}. In particular, we only take posts from the top-50 authors. We also filter out this set for any posts which have been removed, marked by \say{\textit{[removed]}}, as well as the top-writing author, which is simply \say{\textit{[deleted]}}. The final subset consists of 2393 posts from 49 authors.

\paragraph{Trustpilot Reviews.}
We use a subset of 29,490 reviews from the Trustpilot platform, made available by \citet{10.1145/2736277.2741141}. In particular, we use a 10\% sample from the US version of the platform. Each review is marked with the gender (male/female) of the review author, thus creating an adversarial binary classification task. Additionally, the reviews are marked as positive or negative, allowing for a sentiment analysis (utility) task.

\paragraph{Yelp Reviews.}
Finally, we use an authorship identification dataset based on the Yelp reviews dataset, as made available by \citet{utpala-etal-2023-locally}. The dataset features 17,295 reviews written by 10 authors. The reviews are also marked with sentiment, allowing for a downstream utility evaluation.

\subsection{Privatization Procedure}
\label{sec:procedure}
\paragraph{Selected SOTA Methods.}
We evaluate our \textsc{DP-ST} method with three cluster size parameters, namely $k \in \{50000, 100000, 200000\}$. We also use two LLM variants from document reconstruction: \textsc{Llama-3.2-1B-Instruct} and \textsc{Llama-3.2-3B-Instruct} \cite{grattafiori2024llama3herdmodels}. We compare these variants to five recent local DP text privatization methods, ranging from word- to document-level. They are briefly introduced below.

\paragraph{\textsc{TEM} \cite{carvalho2023tem}:} a word-level MDP method leveraging a \textit{truncated} EM, allowing for higher utility word replacements.

\paragraph{\textsc{DP-BART} \cite{igamberdiev-habernal-2023-dp}:} a document-level DP text rewriting method that leverages the BART autoencoder model \cite{lewis-etal-2020-bart}, adding DP noise to the latent representation. We use the \textbf{DP-BART-CLV} variant and a clipping range of (-0.1, 0.1). We test this method using both \textsc{BART-base} and \textsc{BART-large}.

\paragraph{\textsc{DP-Prompt} \cite{utpala-etal-2023-locally}:} a method proposing the use of temperature sampling as an EM equivalent, in order to provide token-level DP guarantees during text generation (here, proxied by a text paraphrasing prompt). Following the original work, we leverage the open-source \textsc{flan-t5} models \cite{chung} as the underlying LLM, specifically the \textsc{large} and \textsc{XL} versions, which are approximately equivalent in size to our selected \textsc{Llama} 1B and 3B models, respectively. As the logit values for each model must be clipped (bounded) to fulfill DP constraints, we clip logits to (\textit{mean}, \textit{mean}$+ 4*$\textit{std}), after measuring all logit values from running the respective models on a random 10 texts from each of our five datasets. These values are provided in our code.

\paragraph{\textsc{DP-MLM} \cite{meisenbacher-etal-2024-dp}:} a method using the EM to perform token replacements by leveraging Masked Language Models. We follow the original implementation, using a \textsc{roberta-base} model \cite{DBLP:journals/corr/abs-1907-11692}.

\paragraph{Setting Document-level Budgets.}
We take a number of steps to ensure a fair comparison between our own method and the selected comparative methods. This is especially important as the methods operate on varying linguistic units. As a solution to this, we decided to fix \textit{document-level} privacy budgets, which are calculated based on the average word count of a given dataset.

We first establish \textit{base} $\varepsilon$ values, chosen to be $\varepsilon \in \{0.1, 0.5, 1\}$. Then, we calculate the per-document budget by scaling (multiplying) the base values by the average word count of a dataset, taking the nearest integer. While we only report results using the base $\varepsilon$ values, the complete set of budgets used is included in Table \ref{tab:budgets} of Appendix \ref{sec:repro}.

Given each per-document budget, a text is privatized by equally distributing the budget based on how a mechanism operates. For example, \textsc{TEM} privatizes each word in a document with $\varepsilon = $ \textit{total budget} / \textit{number of words}, whereas \textsc{DP-BART} simply uses the per-document budget. In the case of \textsc{DP-Prompt}, we enforce a max generation length to that of the original input text, and follow a distribution procedure as with \textsc{TEM} or \textsc{DP-MLM}.

Given the above-outlined procedure, we proceed to privatize all documents in each dataset, for each of the mechanisms with three $\varepsilon$ values. In total, also considering the different model sizes used for some methods, this results in \textbf{180} privatized datasets \footnote{12 method variants $\times$ 5 datasets $\times$ 3 $\varepsilon$ values}.

\subsection{Evaluation Metrics}
We evaluate the privatization outputs of \textsc{DP-ST}, as well as the compared methods, on four criteria: \textit{coherence and utility preservation}, \textit{semantic similarity}, \textit{empirical privacy}, and \textit{relative gain}. We also conduct an \textit{efficiency} test to measure the runtime required for \textsc{DP-ST} and the compared methods.

\paragraph{Coherence and Utility.}
A primary concern of designing an effective DP privatization mechanism comes with maintaining \textit{utility} for downstream use. In this work, we focus particularly on \textit{coherence}, i.e., whether the outputs are natural and fluent.

In particular, we evaluate coherence using the LLM-as-a-Judge paradigm, specifically \textsc{G-Eval} \cite{liu-etal-2023-g}. Following the documentation of the original work\footnote{\url{https://github.com/confident-ai/deepeval}}, as well as taking motivation from related works that use \textsc{G-Eval} for the evaluation of Natural Language Generation tasks \cite{song-etal-2024-finesure,afzal-etal-2024-adapteval}, we define five metrics to comprise the \textit{coherence} evaluation: \textit{Fluency}, \textit{Consistency}, \textit{Clarity}, \textit{Conciseness}, and \textit{Repetitiveness}. We use the \textsc{deepeval} software to operationalize these metrics, firstly by providing a one-sentence \textit{criteria} guideline (found in Appendix \ref{sec:criteria}), which is then converted into \textit{evaluation steps} by \textsc{deepeval}. Each metric is given as a score on the scale of 0-1, and we use OpenAI's \textsc{GPT-4o mini} (2024-07-18) as the evaluator.

In addition, we evaluate downstream utility for the Trustpilot and Yelp datasets, since both datasets are labeled with sentiment scores. For all privatized variants of the dataset, we fine-tune a \textsc{deberta-v3-base} \cite{he2021deberta} model for one epoch, using a 90/10 train/val split. The micro-F1 score on the val split is recorded as the utility score, which can be compared against the baseline of fine-tuning on the original dataset. We provide more details on our fine-tuning procedures in Appendix \ref{sec:repro}.

\paragraph{Semantic Similarity.}
We evaluate to what degree a DP privatization mechanism is capable of preserving the semantic meaning of the original text counterparts. We calculate the pairwise cosine similarity between the embeddings of all private texts to their original counterparts. This is performed using sentence transformers \cite{reimers-gurevych-2019-sentence}, namely the \textsc{all-MiniLM-L12-v2} (ibid), \textsc{all-mpnet-base-v2} (ibid), and \textsc{gte-small} \cite{li2023generaltextembeddingsmultistage} models. The similarity scores for the three models are averaged for each text pair, and then the mean score over each dataset is calculated.

\paragraph{Empirical Privacy.}
Although we have designed the experiments such that all methods privatize a given document with the same \textit{theoretical} DP guarantee, previous works have shown the importance of also measuring \textit{empirical} privacy  \cite{mattern-etal-2022-limits,utpala-etal-2023-locally,meisenbacher-matthes-2024-thinking}, as these can differ from method to method. As such, we test privacy preservation by leveraging the adversarial task associated with each dataset, as introduced in Section \ref{sec:datasets}.

For each private dataset, we train an adversarial \textsc{deberta-v3-base} \cite{he2021deberta} model for one epoch, with the target as the corresponding protected attribute (i.e., author ID or gender). We then compare the performance on the 10\% private validation split, and the reduction in F1, as compared to training a model on the original data, represents the empirical privacy gain afforded by DP privatization. In this process, we distinguish between a \textit{static} adversary, who trains on the original train split, and the \textit{adaptive} attacker, who trains on the train split of the private dataset \cite{mattern-etal-2022-limits}. 

\begin{figure*}[ht!]
    \centering
    \begin{subfigure}[b]{0.46\textwidth}
        \centering
        \includegraphics[width=\textwidth]{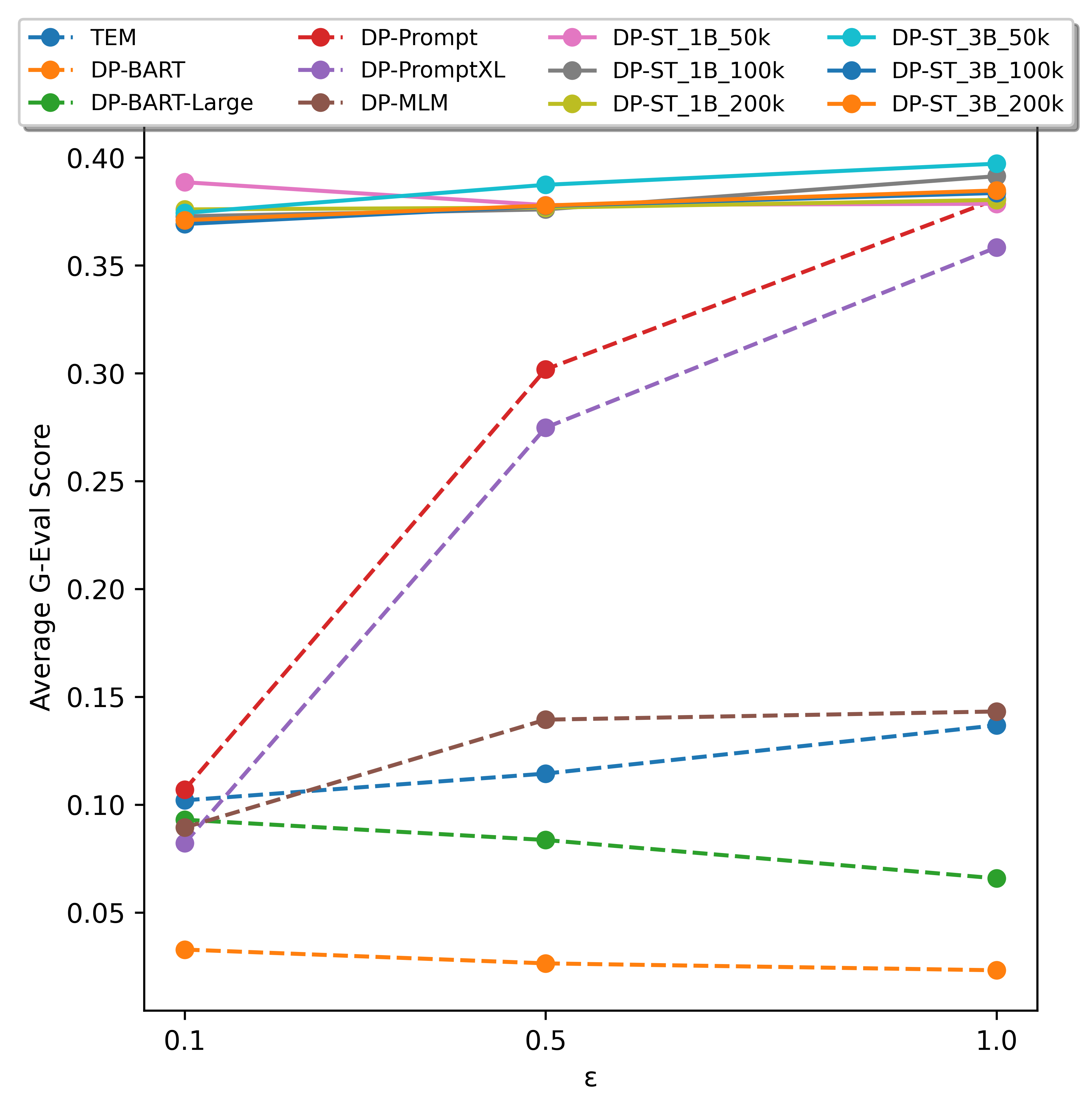}
        \caption{G-Eval (Coherence).}
        \vspace{2pt}
    \end{subfigure}
    \hfill
    \begin{subfigure}[b]{0.46\textwidth}  
        \centering 
        \includegraphics[width=\textwidth]{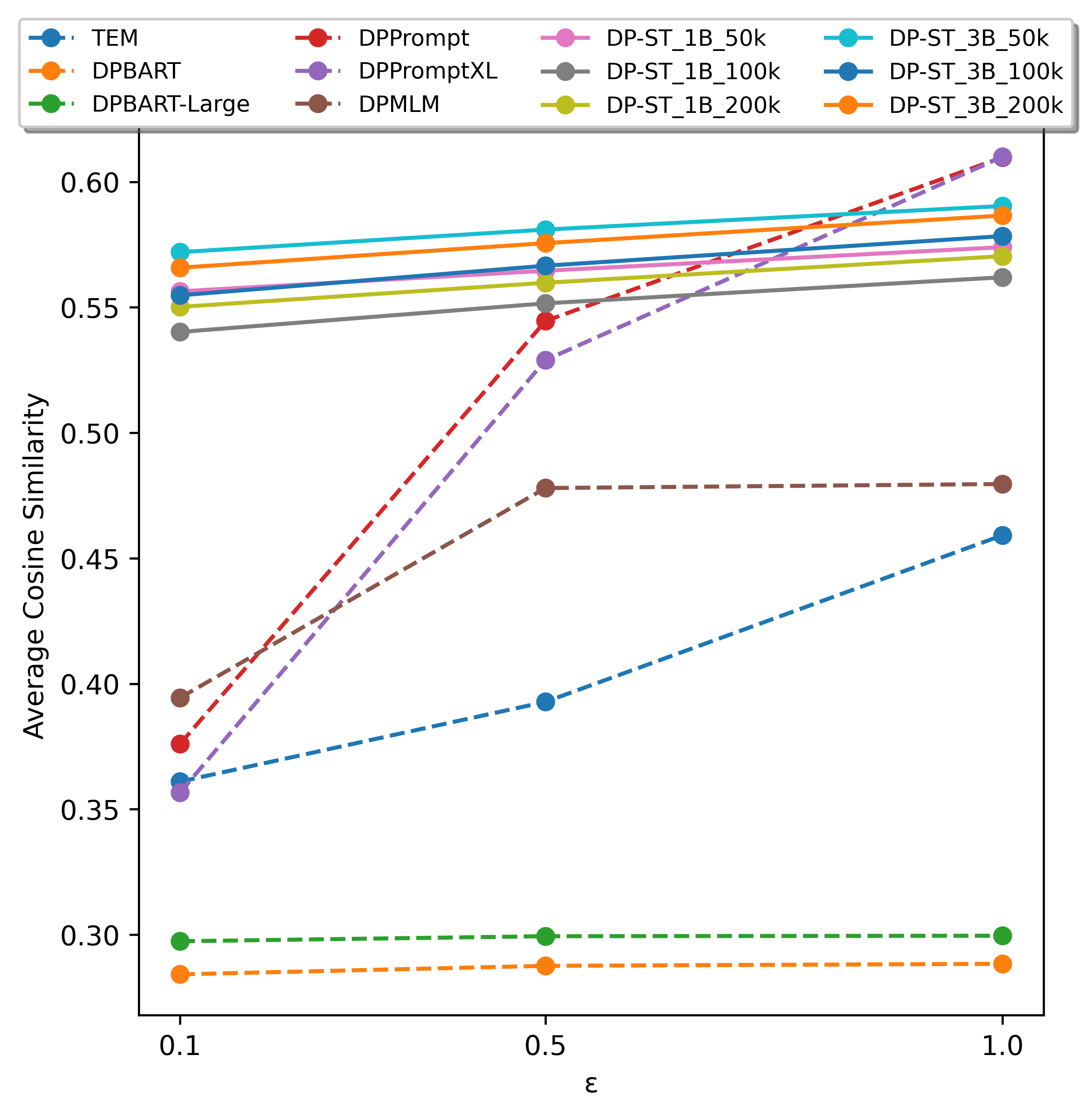}
        \caption{Cosine Similarity.}
        \vspace{2pt}
    \end{subfigure}
    \begin{subfigure}[b]{0.46\textwidth}   
        \centering 
        \includegraphics[width=\textwidth]{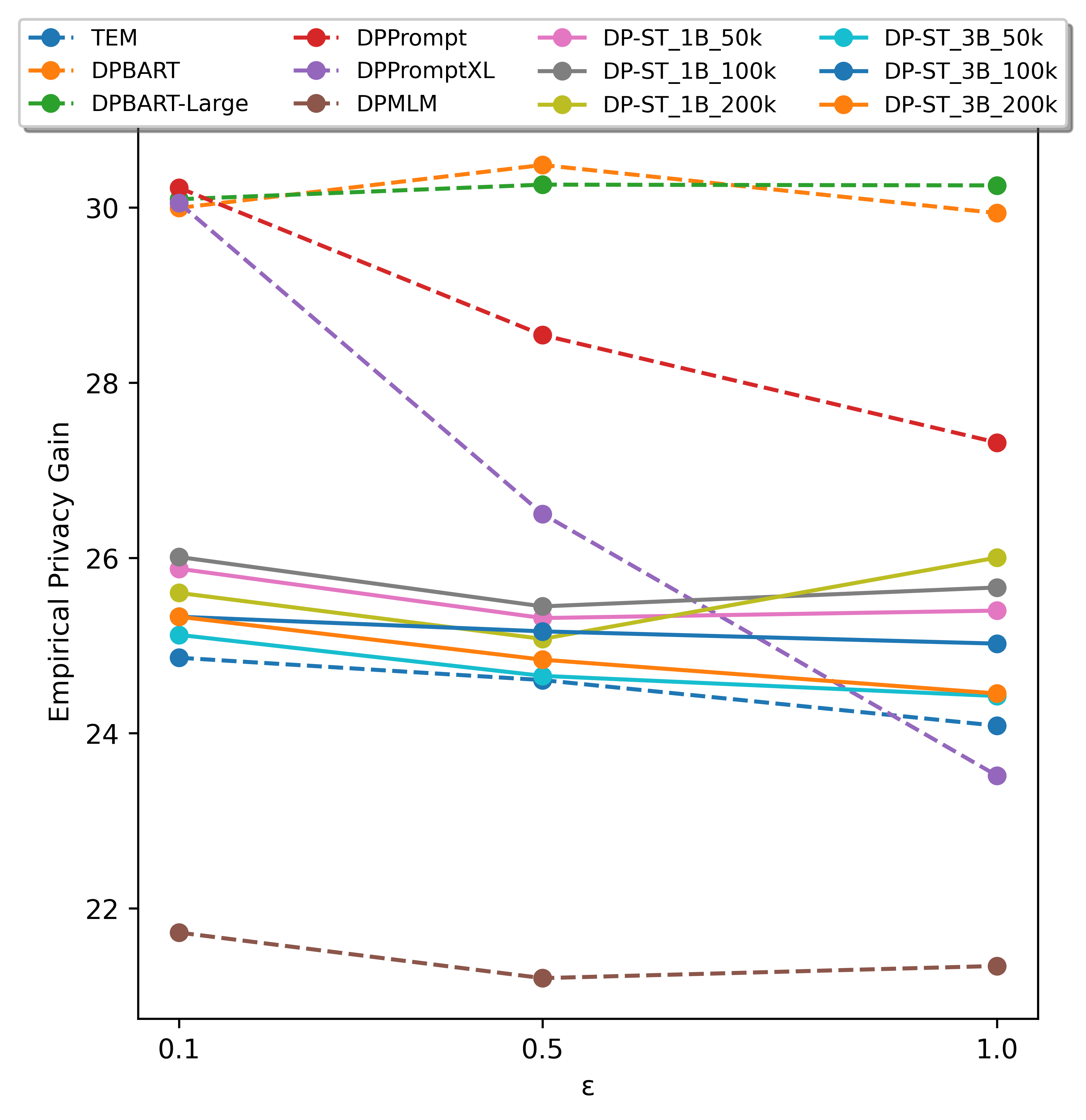}
        \caption{Empirical Privacy.}
    \end{subfigure}
    \hfill
    \begin{subfigure}[b]{0.46\textwidth}   
        \centering 
        \includegraphics[width=\textwidth]{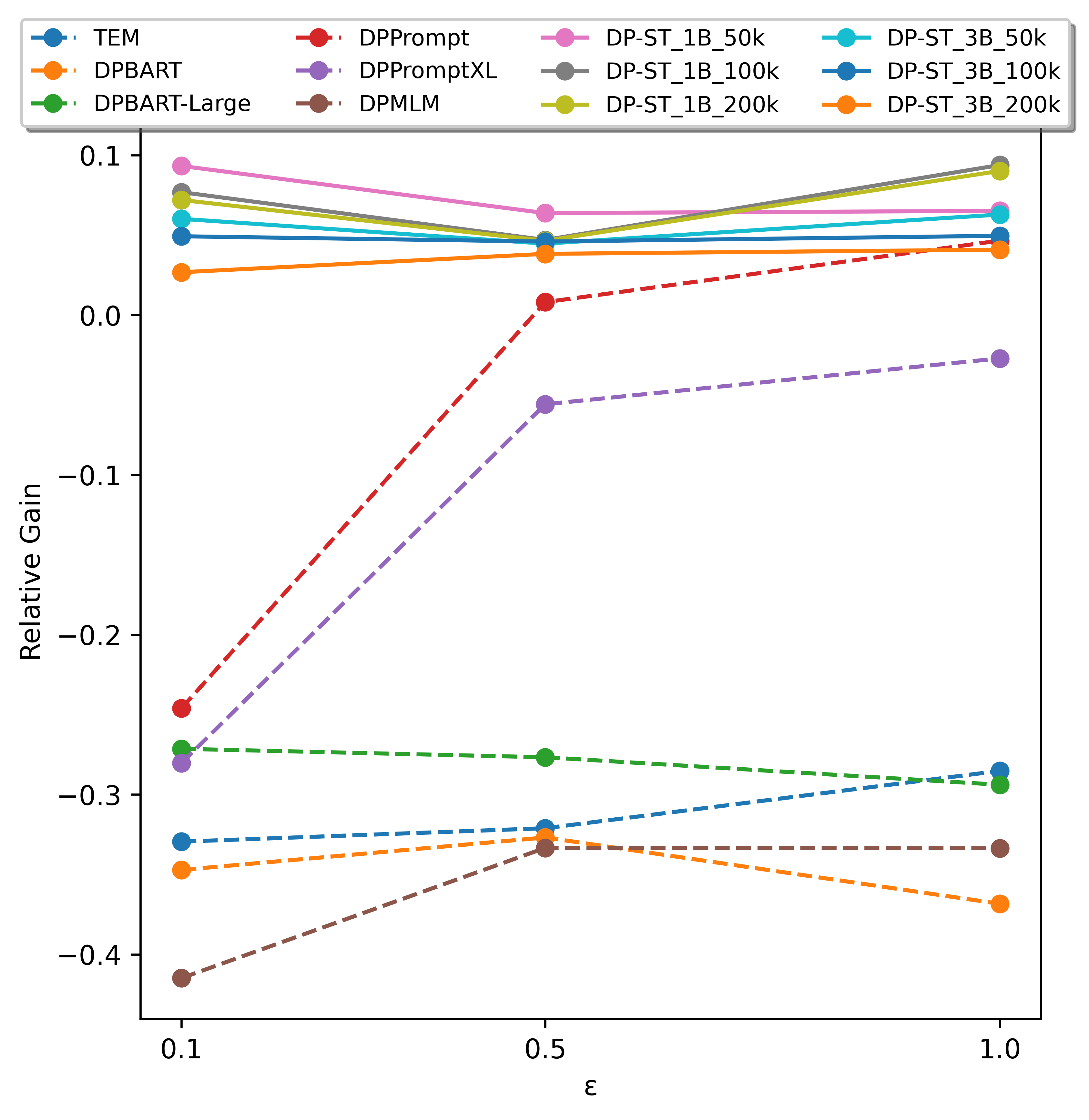}
        \caption{Relative Gain.}  
    \end{subfigure}
    \caption{Aggregated experiment results. All figures portray the average scores over our five datasets. Solid lines indicate \textsc{DP-ST} variants, whereas dashed lines indicate our chosen methods for comparison. In all cases, higher scores represent better results. \textit{Empirical Privacy Gain} measures the average reduction in adversarial performance. For this and Relative Gain, the depicted scores represent the average between the static and adaptive settings.} 
    \label{fig:results}
\end{figure*}

\paragraph{Relative Gain.}
Calculating the \textit{relative gain} (RG) of DP text privatization quantifies the observed benefit of privatization with respect to utility loss. As such, a positive RG denotes that the privacy gains outweigh the utility losses, and a negative score implies the opposite.

We define $\mathcal{U}_o$ to be the baseline utility score, which is represented as the average \textsc{G-Eval} score (i.e., mean over the five metrics), or in the case of Trustpilot and Yelp, the simple mean of the average G-Eval score and the baseline utility task score. We also define $\mathcal{U}_p$ to be the same utility score measured on the privatized datasets. Similarly, we define $EP_o$ and $EP_p$ as the empirical privacy scores, namely the adversarial performance on the baseline and privatized datasets, respectively. We thus define RG as: $RG = \frac{\mathcal{U}_p}{\mathcal{U}_o} - \frac{EP_p}{EP_o}$

RG is therefore maximized by maximizing the left-hand side (minimal utility loss) and minimizing the right-hand side (maximal adversarial performance reduction). We note that in the case of $\mathcal{U}$ calculation for Trustpilot and Yelp, we account for the highly imbalanced datasets (towards positive reviews) by considering the utility change \textit{over majority-class guessing} (MG), thus yielding $\mathcal{U} = \mathcal{U}_{observed} - \mathcal{U}_{MG}$, for both $\mathcal{U}_o$ and $\mathcal{U}_p$.

\paragraph{Efficiency.}
We calculate the runtime required for all of the \textsc{DP-ST} variants and the selected comparison methods. To do this, we choose a random sample (seed 42) of 20 text documents from each of our five datasets, creating a test dataset of 100 documents. We privatize these documents with the six \textsc{DP-ST} configurations and the six comparative baselines, using the dataset-specific $\varepsilon$ values, i.e., three values per dataset (see Section \ref{sec:procedure}). We do not distinguish timing measurements between $\varepsilon$ values, as initial testing demonstrates that the choice of privacy budget does not affect computation time.

We measure the total duration taken by each method to privatize the set of 100 documents, and we also calculate the average duration \textit{per document} and \textit{per word}. The total word count is calculated using word tokenization with \textsc{NLTK}.

\begin{table}[t!]
\centering
    \resizebox{0.9\linewidth}{!}{
\begin{tabular}{l|c|c|c}
\multicolumn{1}{c|}{} & Duration & Avg./Doc. & Avg./Word \\ \hline
\textsc{TEM} & 7022.95 & 23.41 & 0.118 \\
\textsc{DP-BART} (Base) & 491.21 & 1.66 & 0.008 \\
\textsc{DP-BART} (Large) & 738.31 & 2.46 & 0.012 \\
\textsc{DP-Prompt} (Large) & 531.38 & 1.77 & 0.009 \\
\textsc{DP-Prompt} (XL) & 654.09 & 2.18 & 0.011 \\
\textsc{DP-MLM} & 835.28 & 2.78 & 0.014 \\ \hline \hline
\textsc{DP-ST} (1B, 50k) & 735.33 & 2.45 & 0.012 \\ 
\textsc{DP-ST} (1B, 100k) & 761.99 & 2.54 & 0.013 \\ 
\textsc{DP-ST} (1B, 200k) & 631.66 & 2.11 & 0.011 \\ 
\textsc{DP-ST} (3B, 50k) & 1127.49 & 3.76 & 0.019 \\ 
\textsc{DP-ST} (3B, 100k) & 1220.01 & 4.07 & 0.020 \\ 
\textsc{DP-ST} (3B, 200k) & 986.54 & 3.29 & 0.017 \\
\end{tabular}
}
\caption{Efficiency test results. Values are in seconds, and they represent the (average) duration taken for each method to run over the selected set of 100 documents and three $\varepsilon$ values (thus, 300 documents total).}
\label{tab:time}
\end{table}

\subsection{Results}
Figure \ref{fig:results} illustrates the results for coherence, similarity, privacy, and relative gain, shown as the average scores for each method over all five datasets. The complete results are located in Appendix \ref{sec:tables}. The efficiency test results are found in Table \ref{tab:time}.

\section{Discussion}
\label{sec:discuss}
\paragraph{The strengths of \textsc{DP-ST}.}
We find that an immediate strength of \textsc{DP-ST} comes with its demonstrated ability to produce private output texts that are clear, concise, and natural. This can be attributed to the LLM post-processing step, suggesting that such a step could also improve the outputs of other methods. This boost in coherence comes in tandem with strong semantic preservation (cosine similarity), as well as acceptable empirical privacy gains. Most importantly, \textsc{DP-ST} is the only tested method that can, on average, maintain positive relative gains, an important result in the discussion of the merits of DP text privatization. Another interesting trend that can be observed is the general robustness of \textsc{DP-ST} over $\varepsilon$ values, where the drop in performance is not as steep as $\varepsilon$ increases.

These strengths come at comparable computation times to state-of-the-art methods (Table \ref{tab:time}), especially when considering the multi-stage process of \textsc{DP-ST} (decomposition - privatization - reconstruction). Particularly when using a smaller LLM (e.g., \textsc{Llama-3.2-1B}), we find near comparable runtimes to all selected baselines, and given the improved coherence, semantic similarity, and trade-offs, we argue that this comparability in efficiency is a clear additional strength of \textsc{DP-ST}.

\begin{table*}[t!]
\small
\centering
\begin{subtable}[t]{0.99\linewidth}
    \small
     \centering
     \resizebox{0.99\linewidth}{!}{
\begin{tabular}{l|p{0.99\textwidth}}
Original Text & \textit{ok so in one weird moment 2 years ago i was angry, it caused an arousal, and then a weird bad thought popped into my head about harming the person making me angry. Back then i didnt have the ocd theme that causes such things, i just was like "ugh no! why did i even have this thought?!" and just ignored it (guess its called the ol\' good days!) i didnt get stuck in it, (...) is it like my mind throwing something random and all? or did my mind just crash because of the different kinds of feelings that were mixed together weirdly} \\ \hline
Extracted Triples & [`i | was | so angry', `weird thought | popped into | my head', `i | did even have | thought', `i | get | stuck in it', `my mind | is | like throwing', `my mind | throwing | something'] \\ \hline
Private Triples &  [`I | was more angry than | scared', `ideas | are becoming in | my mind', `i | realized | I was going', `I | was dragged For | now', `my mind | Immediately adds to | puzzle', `my mind | flashed For | moment'] \\ \hline
Private Text & \textit{I was more angry than scared, and my mind is becoming increasingly filled with my ideas, which immediately adds to my puzzle, and I realized I was going somewhere, but now I'm being dragged there, and my mind flashed that moment.}
\end{tabular}
}
\caption{Example from Reddit Mental Health.}
\end{subtable}     
    \vspace{5pt}
    
    \begin{subtable}[t]{0.99\linewidth}
    \small
     \centering
\resizebox{0.99\linewidth}{!}{
\begin{tabular}{l|p{0.99\textwidth}}
Original Text & \textit{Enjoyed the experience!: I thoroughly enjoyed the experience of creating my own photo album and adding my own narratives. I now have 6 of the Crewe photo books and I am really proud of them. The price is competitive and delivery and quality excellent. Will definitely use again (can't wait!) and would highly recommend the service to anyone. Well done Jessops} \\ \hline
Extracted Triples & [`I | enjoyed | experience', `my own photo album | adding | my own narratives', `I | am | really proud'] \\ \hline
Private Triples & [`It | was great experience for | young person', `pictures | now is considered in | your pictures gallery', `i | am proud for | this'] \\ \hline
Private Text & \textit{It was a great experience for young person and is now considered part of your pictures gallery, and I am proud of this. }\\ 
\end{tabular}
}
\caption{Example from Trustpilot.}
\end{subtable}  
\caption{Output examples from \textsc{DP-ST} (3B,200k) on texts from Reddit Mental Health and Trustpilot. The order of the \textit{Private Triples} match that of the \textit{Extracted Triples} (from OpenIE), showing the one-to-one DP outputs. \textit{Private Text} represents the output from the LLM post-processing step. Example (a) has been truncated for readability.}
\label{tab:output_examples}
\end{table*}

\paragraph{The limitations of \textsc{DP-ST}.}
We acknowledge the trade-offs that come with these strengths, which are ultimately rooted in the semantic decomposition stage of \textsc{DP-ST}, where an input text is reduced to its component triples. While this highlights the semantic core of the text, it also discards nuanced information about the writing style, important attributes, or modifiers outside of the triples. Preserving this information, although presumably disadvantageous from a privacy perspective, would undoubtedly lead to semantically richer output texts.

Examples of the transformation of an input text to triples to a private output document can be found in Table \ref{tab:output_examples}. As can be seen, while \textsc{DP-ST}, namely the triple extraction stage, is able to extract core semantic triples from the text, much of the surrounding context is ignored. This typically results in a significantly \say{distilled} output text -- something which certainly leads to higher privacy, but at the cost of less expressive language. We believe that can also be largely attributed to the use of OpenIE, which is limited in its extractive capabilities. The upshot of this is the fact that \textsc{DP-ST} is modular, and future improvements can conveniently swap in more capable triple extraction techniques.

\paragraph{What is being privatized?}
An important consideration to make is the question of what exactly \textsc{DP-ST} is privatizing, and the implications of this decision. The recent literature has introduced a diversity of ways in which a full text can be made private under DP, yet these methods differ significantly in their underlying design and resulting privacy guarantees. While some works focus on providing document-level guarantees via a single DP mechanism run, we opt for the \textit{divide-and-conquer} paradigm, which allows for greater flexibility and higher granularity, limiting our privacy focus to the \textit{semantic core} rather than the full text. We stress that this guarantee only holds for our \textit{privatization neighborhoods}, or semantic clusters of triples.

To address these complexities, we design an experimental setup in which all methods, regardless of DP notion, are tested under equal \textit{document}-level privacy budgets. This, however, cannot account for the differences in guarantees offered, for example, between $\varepsilon$-DP and MDP. Nevertheless, we pose that although not all DP methods are made equally, we can at least gain a sense of their relative performance when enforcing uniform budgets.

\paragraph{Focusing on the trade-off.}
In our evaluation, we focus on the trade-off between text coherence / downstream utility (measured by \textsc{G-Eval} and task performance) and empirical privacy. We find that all compared methods struggle at the chosen $\varepsilon$ levels to achieve a positive balance, suggesting that in these cases, DP privatization does not carry an advantage. Only \textsc{DP-ST} achieves positive gains on average, and we observe that this generally comes as a result of balancing strong coherence with slightly lower privacy protections. Conversely, we see the effect of imbalanced privatization, where low coherence and semantic similarity lead to high privacy protections, but unsatisfactory trade-offs, such as with \textsc{TEM}, \textsc{DP-BART}, or \textsc{DP-MLM}.

This becomes an interesting point of investigation for future work, namely, to design DP privatizations that can find acceptable privacy-utility trade-offs even at stricter privacy budgets. Of course, this comes with the consideration of how exactly the trade-off is quantified, and the potentially greater importance of privacy over utility, or vice versa.

\paragraph{The effect of LLM parameters and cluster size.}
We analyze the results from our two selected reconstruction LLMs (\textsc{LLaMa-3.2-1B} and \textsc{LLaMa-3.2-3B}), in conjunction with the three tested cluster sizes (50k, 100k, 200k). From Figure \ref{fig:results}, one can see that the 3B model outperforms the 1B model in semantic similarity, whereas the distinction for coherence scores is not so clear. However, the 1B model clearly performs better in terms of empirical privacy, which leads to better overall relative gains. This suggests a potential paradox in using \textsc{DP-ST}, where using larger models may improve the semantic resemblance of the private outputs to the original outputs, but in doing so, may weaken the resulting privacy-utility trade-offs.

The story with the cluster size is more complex. One clear trend is that of relative gains: our results imply that the smaller the number of clusters (i.e., the greater the intra-cluster \say{privatization range}), the better the trade-off. This, however, is not universally true, and seemingly depends also on the privacy budget. Complexities arise as sometimes 100k outperforms 200k \textit{and} 50k, such as in the lower privacy budgets with the empirical privacy results. As such, we introduce another privatization parameter that, on one side, may introduce complexity to privatization, but on the other hand, offers greater flexibility and the ability to tune the privacy-utility trade-off. \textsc{DP-ST} enables a variable cluster size parameter, which must only be preceded by re-running clustering on the triple corpus.

\section{Conclusion}
We introduce \textsc{DP-ST}, a DP privatization method that leverages semantic triple extraction for private document generation under local DP guarantees. We show the effectiveness of \textsc{DP-ST}, particularly when equipped with LLM post-processing, in creating coherent, yet privacy-preserving text outputs, ultimately finding a superior balance than other DP methods. We achieve such results at relatively low privacy budgets, providing a way forward for reasonable DP privatization under the \textit{divide-and-conquer} paradigm. Accordingly, we define two points for future work: (1) continued studies on the feasibility of the \textit{divide-and-conquer} paradigm, and (2) increased work on post-processing techniques for supporting DP text privatization.

\newpage
\section*{Limitations}
We acknowledge the main limitations of our work, which firstly pertain to the reliance of our \textsc{DP-ST} method on extracted semantic triples from a public data corpus (i.e., FineWeb). We decided to stop the extraction process after 15 million triples, which could potentially limit the semantic expressiveness of the created triple corpus. Additionally, we utilize the OpenIE tool for efficient triple extraction, but this method does not represent the SOTA and is limited in its extractive capabilities. Future improvements of \textsc{DP-ST}, and the triple corpus creation, should focus on the usage of more capable extraction methods. The final limitation related to \textsc{DP-ST} is the built-in fallback option when no triples are extracted from an input text, simply returning the input without modifications. This may have negative outcomes regarding empirical privacy, especially when privatizing shorter texts.

We caution that the experiment results between DP methods are not directly comparable, as we test MDP-based methods (\textsc{TEM}), $\varepsilon$-DP methods (\textsc{DP-Prompt}, \textsc{DP-MLM}), and $(\varepsilon, \delta)$-DP-based methods (\textsc{DP-BART}), all in the local setting. Our \textsc{DP-ST} method also operates on the local DP setting, with $\varepsilon$-DP guarantees; however, we leverage a relaxed \textit{neighborhood-aware} local DP notion based on \textit{privatization neighborhoods}. Thus, while the document-level budgets were made uniform with respect to $\varepsilon$, there are deeper considerations to be made when weighing the true comparability between various DP text privatization methods. 

Another important limitation with our evaluation procedure concerns the potential effect of \textit{data contamination}, namely, if the LLMs we used for document reconstruction have seen any of the evaluation datasets during their pre-training. However, we believe this risk to be minimal, as the original texts are never seen by the LLMs, but rather only a collection of private triples derived from these original texts. As such, the original texts presumably have minimal influence on the final private output, i.e., the reconstructed text from the LLM.

Finally, while we conduct an initial discussion on the effect of model size (i.e., for LLM document reconstruction) and cluster size, we do not perform a rigorous analysis of the effects of these design choices. Future work would serve the \textsc{DP-ST} method well in guiding a more informed understanding of how these choices affect the utility and privacy preservation capabilities of \textsc{DP-ST}.

\section*{Ethics Statement}
Our use of public datasets for our privacy evaluations, namely for adversarial tasks, is not aligned with their initial intended usage. We believe such concerns to be mitigated, as no personally-identifiable information is contained within. Similarly, the use of FineWeb for the public triple corpus creation may leak undetected personal information that evaded filtering originally. We did not perform any filtering beyond that described in this work.


\bibliography{custom}

\appendix

\section{Extended Proof for DP-ST}
\label{sec:proof-appendix}
Suppose $\mathcal{X}$ is the set of all the triples in the universe. Let $\mathcal{P}$ be the set of all triples from a sensitive text and let the set of all triples derived from public texts be $\mathcal{V}$, which is further segmented into $n$ mutually exclusive clusters or subsets, each denoted by $\mathcal{C}_i$. It can happen that there is an overlap in the sets $\mathcal{P}$ and $\mathcal{V}$. Our DP mechanism $\mathcal{M}: \mathcal{P} \to \mathcal{V}$ takes a triple $x \in \mathcal{P}$ derived from a sensitive text and returns its privatized version from the cluster $\mathcal{C}_i$ in which the embedding of the sensitive triple is located.

The Exponential Mechanism \cite{4389483} is used to select a privatized triple using a utility function $u: \mathcal{P} \times \mathcal{V} \to \mathbb{R}$ that maps sensitive and private triple pairs ($x$, $y$) to a possibility score. Considering $\Phi: \mathcal{X} \to \mathbb{R}^d$ as the embedding function mapping a triple to a $d$-dimensional Euclidean space, our utility function is defined as:
\begin{displaymath}
    u(x, y)  =  \cos\left(\Phi(x), \Phi(y)\right)
\end{displaymath}
Since $\cos$ is bounded between -1 and 1, the $l_2$-sensitivity of our utility function, $\Delta u$, is 2. The sensitivity is further reduced in our case because of the application of the exponential mechanism in just a single cluster, where embeddings are presumably similar to each other. Empirically, we found that the cosine distance between any two vectors in any cluster is always non-negative, further reducing the value of sensitivity to 1. This can also be achieved by bounding the cosine values to 0 and 1. This is not required and is an innate behavior because of the high number of clusters in our case. However, we include an explicit check for this in our code, where negative values are changed to 0.

\subsection{Proof}
The EM outputs the privatized version of the sensitive triple $x$ by selecting a triple $z \in \mathcal{V}$ with a probability proportional to $\exp \left( \frac{\varepsilon u(x, z)}{2\Delta u} \right)$. Hence, for all cluster $\mathcal{C}_i \subset \mathcal{V}$ that are mutually exclusive to each other, and two triple $x, y \in \mathcal{C}_i$, the ratio of output probability distribution for any output $z$ of our DP mechanism can be given as:

{\scriptsize \begin{align*}
    \frac{Pr[M(x) = z]}{Pr[M(y) = z]} 
    &= \frac{ \left( \frac{\exp \left( \frac{\varepsilon u(x, z)}{2\Delta u} \right)}
                          {\sum_{z_i \in \mathcal{V}} \exp \left( \frac{\varepsilon u(x, z_i)}{2\Delta u} \right)} \right)
            }
            { \left( \frac{\exp \left( \frac{\varepsilon u(y, z)}{2\Delta u} \right)}
                          {\sum_{z_i \in \mathcal{V}} \exp \left( \frac{\varepsilon u(y, z_i)}{2\Delta u} \right)} \right)
            } \\
    &= \left( \frac{\exp \left( \frac{\varepsilon u(x, z)}{2\Delta u} \right)}
                   {\exp \left( \frac{\varepsilon u(y, z)}{2\Delta u} \right)} \right) 
        \cdot
        \left( \frac{\sum_{z_i \in \mathcal{V}} \exp \left( \frac{\varepsilon u(x, z_i)}{2\Delta u} \right)}
                    {\sum_{z_i \in \mathcal{V}} \exp \left( \frac{\varepsilon u(y, z_i)}{2\Delta u} \right)} \right)
\end{align*}}

Since the $l_2$ sensitivity of the utility function is defined as
\begin{align*}
\Delta u = \underset{z \in \mathcal{V}}{\max} \underset{x, y \in \mathcal{P}}{\max} |u(x, z) - u(y^\prime, z)|
\end{align*},
the first term in the above equation is upper bounded by the following:
\begin{align*}
    \frac{\exp \left( \frac{\varepsilon u(x, z)}{2\Delta u} \right)}
                   {\exp \left( \frac{\varepsilon u(y, z)}{2\Delta u} \right)}
    &= \exp\left( \frac{\varepsilon \left( u(x, z) - u(y, z) \right)}{2 \Delta u} \right) \\
    & \leq \exp\left(\frac{\varepsilon}{2}\right)
\end{align*}. Similarly, the second term in the above equation is also upper bounded by the same value and our ratio can be simplified to the following:

\begin{align*}
       \frac{Pr[M(x) = z]}{Pr[M(y) = z]} &\leq \exp\left(\frac{\varepsilon}{2}\right) \cdot \exp\left(\frac{\varepsilon}{2}\right) \\
            &= \exp\left(\varepsilon\right)
\end{align*}

\section{Reconstruction Prompt}
\label{sec:prompt}
In Table \ref{tab:prompt}, we provide the few-shot prompt used with our chosen LLMs for the reconstruction of a set of triples into a coherent document.

\begin{table*}[htbp]
\centering
\footnotesize
\caption{Prompt for LLM Document Reconstruction.}
\begin{tabular}{p{0.95\linewidth}}
\hline
\textbf{Prompt}
\\ \hline
Generate a concise text for the given set of triples. Ensure that the generated output only includes the provided information from the triples, but feel free to fill in the gaps where sensible. If necessary, ignore triples that do not fit into the larger context. It is very important that the output is grammatically correct, natural, and logical. Provide a text that captures the semantic meaning of the triples, without being too verbose or lengthy. Do not provide any further explanation, only provide the output text. \\

\textbf{user:} Input triples: [{`object’: `Mike\_Mularkey’,`property’: `coach’,`subject’: `Tennessee\_Titans’}]\\
\textbf{assistant:} Output text: Mike Mularkey is the coach of the Tennessee Titans.\\
\textbf{user:} Input triples: [{`object’: `Albert\_E.\_Austin’, `property’: `successor’, `subject’: `Alfred\_N.\_Phillips’}, {`object’: `Connecticut’, `property’: `birthPlace’, `subject’: `Alfred\_N.\_Phillips’}, {`object’: `United\_States\_House\_of\_Representatives’, `property’: `office’, `subject’: `Alfred\_N.\_Phillips’}]\\
\textbf{assistant:} Output text: Albert E. Austin succeeded Alfred N. Phillips who was born in Connecticut and worked at the United States House of Representatives. \\
\textbf{user:} Input triples: [{`object’: `College\_of\_William\_\&\_Mary’, `property’: `owner’, `subject’: `Alan\_B.\_Miller\_Hall’}, {`object’: `2009-06-01’, `property’: `completionDate’, `subject’: `Alan\_B.\_Miller\_Hall’}, {`object’: `101 Ukrop Way’, `property’: `address’, `subject’: `Alan\_B.\_Miller\_Hall’}, {`object’: `Williamsburg,\_Virginia’, `property’: `location’, `subject’: `Alan\_B.\_Miller\_Hall’}, {`object’: `Robert\_A.\_M.\_Stern’, `property’: `architect’, `subject’: `Alan\_B.\_Miller\_Hall’}] \\
\textbf{assistant:} Output text: The Alan B Miller Hall’s location is 101 Ukrop Way, Williamsburg, Virginia. It was designed by Robert A.M. Stern and was completed on 1 June 2009. Its owner is the College of William and Mary. \\
\textbf{user:} [PRIVATE TRIPLES] \\
\textbf{assistant:} \\
\hline
\end{tabular}
\label{tab:prompt}
\end{table*}

\section{LLM-as-a-Judge Criteria}
\label{sec:criteria}
For the calculation of the \textit{coherence} scores using LLM-as-a-Judge, we defined the following five metrics, given with their corresponding \textit{criterion}:

\begin{itemize}
    \itemsep 0em
    \item \textbf{Fluency}: \textit{Measures how smoothly the text reads, focusing on grammar and syntax.}
    \item \textbf{Consistency}: \textit{Ensures the text maintains a uniform style and tone throughout.}
    \item \textbf{Clarity}: \textit{Evaluates how easily the actual output can be understood by the reader.}
    \item \textbf{Conciseness}: \textit{Assesses whether the text is free of unnecessary words or details.}
    \item \textbf{Repetitiveness}: \textit{Checks for redundancy or repeated information in the text.}
\end{itemize}

\section{Reproducibility}
\label{sec:repro}
All code required to replicate our experiments are included in our public repository. This excludes the sharing of the Weaviate vector database, as it is very large in size. Instead, code to replicate the creation of the database is provided. In addition, we clarify several important points for reproducibility.

\paragraph{Sampling.}
All random sampling performed in this work was done using a random seed of 42. This comes with the exception of dataset shuffling during the three training repetitions for each fine-tuning procedure, for which no seed was set.

\paragraph{Training Parameters.}
Fine-tuning was performed using \textsc{Trainer} from \textsc{transformers}. Defaults were kept, except for the batch size, which we varied alongside the input size of \textsc{deberta-v3-base}. For Yelp and Reuters, we used an input length of 512 with a batch size of 32, for Trustpilot and Reddit we used an input length of 256 with batch size of 64, and for Spooky Authors we used an input length of 128 for a batch size of 128.

\paragraph{Hardware.}
The entire evaluation procedure was performed using a NVIDIA RTX A6000 GPU (48GB), which was also used for privatization with \textsc{DP-ST}. For the other tested DP methods, we ran these on either a NVIDIA Quadro RTX 8000 (48GB) or NVIDIA Tesla V100 (16GB).

\paragraph{Document-level Privacy Budgets}
In Table \ref{tab:budgets}, we provide all used document-level privacy budgets.

\begin{table}[ht!]
    \centering
    \resizebox{\linewidth}{!}{
\begin{tabular}{lc|ccc}
\multicolumn{2}{l|}{} & \multicolumn{3}{c}{Document Budget ($\varepsilon$)} \\ \hline
\multicolumn{1}{l|}{Dataset} & Avg. Words/Text & \multicolumn{1}{c}{0.1} & \multicolumn{1}{c}{0.5} & \multicolumn{1}{c}{1} \\ \hline
\multicolumn{1}{l|}{Reuters} & 575.21 & 57.5 & 287.5 & 575 \\
\multicolumn{1}{l|}{Spooky} & 30.39 & 3.0 & 15.0 & 30 \\
\multicolumn{1}{l|}{Reddit} & 141.70 & 14.1 & 70.5 & 141 \\
\multicolumn{1}{l|}{Trustpilot} & 59.75 & 5.9 & 29.5 & 59 \\
\multicolumn{1}{l|}{Yelp} & 208.62 & 20.8 & 104.0 & 208
\end{tabular}
}
\caption{Document-level budgets. Given our base $\varepsilon$ values, we scale the allocated overall budget per document based on the average token length of documents.}
\label{tab:budgets}
\end{table}

\paragraph{Relative Gain Calculation.} As previously noted, we adjust the RG scores, specifically the utility portion, by considering utility changes over majority class guessing performance ($\mathcal{U}_{MG}$). For Trustpilot, this is calculated to be 95.83, and for Yelp 96.65. 

\section{Experiment Results}
\label{sec:tables}

In Tables \ref{tab:reuters}, \ref{tab:spooky}, \ref{tab:reddit}, \ref{tab:trustpilot}, and \ref{tab:yelp}, we provide the full results for all conducted experiments in this work. The provided epsilon values (0.1, 0.5, and 1) represent the \textit{base} $\varepsilon$ values. For the exact values used for this dataset, please refer to Table \ref{tab:budgets}. \textit{GE} denotes the average G-Eval score over the five defined metrics, \textit{CS} denotes average cosine similarity, \textit{EP} denotes the empirical privacy results for either the static (s) or adaptive (a) settings, and \textit{RG} denotes the relative gain results. Note that Tables \ref{tab:trustpilot} and \ref{tab:yelp} (Trustpilot and Yelp) also contain the \textit{Util} score, which gives the results for the associated utility tasks. For metric evaluations that require model training, (e.g., utility or adaptive attacker), we provide the average result over three training runs; the standard deviation is also given as a subscript.

\begin{table*}[ht]
\centering
    \resizebox{\linewidth}{!}{
\begin{tabular}{l|cccccc|cccccc|cccccc}
 & \multicolumn{6}{c|}{0.1} & \multicolumn{6}{c|}{0.5} & \multicolumn{6}{c}{1} \\ \cline{2-19} 
 & GE & CS & EP(s) & EP(a) & RG(s) & RG(a) & GE & CS & EP(s) & EP(a) & RG(s) & RG(a) & GE & CS & EP(s) & EP(a) & RG(s) & RG(a) \\ \hline
Baseline & 0.697 & - & 12.35 & 12.35 & - & - & 0.697 & - & 12.35 & 12.35 & - & - & 0.697 & - & 12.35 & 12.35 & - & - \\ \hline
\textsc{TEM} & 0.098 & 0.370 & 1.59 & $1.46_{0.7}$ & 0.012 & 0.022 & 0.103 & 0.383 & 1.20 & $1.73_{0.7}$ & 0.051 & 0.008 & 0.112 & 0.407 & 1.59 & $1.20_{0.6}$ & 0.032 & 0.064 \\
\textsc{DP-BART} (Base) &  0.026 & 0.286 & 0.80 & $2.12_{0.2}$ & -0.027 & -0.134 & 0.011 & 0.290 & 1.20 & $1.73_{0.4}$ & -0.081 & -0.124 & 0.011 & 0.285 & 2.39 & $1.86_{0.2}$ & -0.178 & -0.135 \\
\textsc{DP-BART} (Large) & 0.097 & 0.295 & 2.79 & $1.73_{0.7}$ & -0.087 & -0.001 & 0.074 & 0.301 & 1.20 & $1.46_{0.5}$ & 0.009 & -0.012 & 0.024 & 0.298 & 1.20 & $1.86_{0.5}$ & -0.063 & -0.116 \\
\textsc{DP-Prompt} (Large) & 0.349 & 0.574 & 2.39 & $2.12_{0.5}$ & 0.307 & 0.329 & 0.431 & 0.651 & 1.99 & $2.52_{0.8}$ & 0.457 & 0.414 & 0.439 & 0.650 & 1.99 & $2.66_{0.7}$ & 0.469 & 0.414 \\
\textsc{DP-Prompt} (XL) & 0.224 & 0.491 & 1.99 & $2.26_{0.4}$ & 0.160 & 0.138 & 0.423 & 0.623 & 1.20 & $2.39_{0.3}$ & 0.510 & 0.413 & 0.424 & 0.620 & 1.59 & $2.12_{0.4}$ & 0.480 & 0.437 \\
\textsc{DP-MLM} & 0.050 & 0.335 & 0.40 & $2.39_{0.3}$ & 0.039 & -0.122 & 0.056 & 0.338 & 0.80 & $1.73_{0.7}$ & 0.016 & -0.060 & 0.058 & 0.344 & 1.20 & $2.26_{0.2}$ & -0.014 & -0.100 \\ \hline \hline
\textsc{DP-ST} (1B, 50k) & 0.367 & 0.562 & 2.39 & $1.86_{0.8}$ & 0.333 & 0.376 & 0.360 & 0.570 & 3.98 & $2.92_{1.1}$ & 0.194 & 0.280 & 0.361 & 0.583 & 5.98 & $2.12_{0.8}$ & 0.034 & 0.346 \\
\textsc{DP-ST} (1B, 100k) & 0.345 & 0.538 & 2.39 & $1.86_{0.2}$ & 0.301 & 0.344 & 0.356 & 0.554 & 5.18 & $2.66_{0.5}$ & 0.091 & 0.295 & 0.372 & 0.567 & 4.38 & $1.73_{0.7}$ & 0.179 & 0.394 \\
\textsc{DP-ST} (1B, 200k) & 0.343 & 0.555 & 5.98 & $1.73_{0.4}$ & 0.008 & 0.352 & 0.332 & 0.563 & 5.58 & $1.99_{0.6}$ & 0.025 & 0.315 & 0.370 & 0.581 & 2.79 & $1.86_{0.4}$ & 0.305 & 0.380 \\
\textsc{DP-ST} (3B, 50k) & 0.352 & 0.584 & 3.19 & $2.26_{1.0}$ & 0.247 & 0.322 & 0.355 & 0.594 & 4.78 & $2.66_{2.1}$ & 0.122 & 0.294 & 0.385 & 0.608 & 4.38 & $2.26_{0.4}$ & 0.198 & 0.369 \\
\textsc{DP-ST} (3B, 100k) & 0.352 & 0.559 & 3.59 & $2.39_{1.5}$ & 0.214 & 0.312 & 0.360 & 0.578 & 4.78 & $1.86_{0.8}$ & 0.129 & 0.366 & 0.369 & 0.591 & 5.58 & $1.33_{0.2}$ & 0.078 & 0.422 \\
\textsc{DP-ST} (3B, 200k) & 0.342 & 0.576 & 4.38 & $2.66_{0.2}$ & 0.136 & 0.275 & 0.360 & 0.591 & 5.58 & $2.39_{0.3}$ & 0.065 & 0.323 & 0.375 & 0.606 & 3.98 & $2.52_{1.0}$ & 0.216 & 0.334
\end{tabular}
}
\caption{Full results for the Reuters dataset.}
\label{tab:reuters}
\end{table*}

\begin{table*}[ht]
\centering
    \resizebox{\linewidth}{!}{
\begin{tabular}{l|cccccc|cccccc|cccccc}
 & \multicolumn{6}{c|}{0.1} & \multicolumn{6}{c|}{0.5} & \multicolumn{6}{c}{1} \\ \cline{2-19} 
 & GE & CS & EP(s) & EP(a) & RG(s) & RG(a) & GE & CS & EP(s) & EP(a) & RG(s) & RG(a) & GE & CS & EP(s) & EP(a) & RG(s) & RG(a) \\ \hline
Baseline & 0.538 & - & 83.15 & 83.15 & - & - & 0.538 & - & 83.15 & 83.15 & - & - & 0.538 & - & 83.15 & 83.15 & - & - \\ \hline
\textsc{TEM} & 0.110 & 0.377 & 33.96 & $47.14_{0.7}$ & -0.204 & -0.362 & 0.113 & 0.398 & 35.65 & $48.20_{1.1}$ & -0.219 & -0.370 & 0.151 & 0.478 & 39.94 & $51.46_{1.6}$ & -0.200 & -0.338 \\
\textsc{DP-BART} (Base) & 0.034 & 0.297 & 39.79 & $39.68_{0.0}$ & -0.415 & -0.414 & 0.038 & 0.298 & 39.68 & $39.68_{0.0}$ & -0.407 & -0.407 & 0.034 & 0.300 & 39.43 & $39.68_{0.0}$ & -0.411 & -0.414 \\
\textsc{DP-BART} (Large) & 0.101 & 0.312 & 34.88 & $39.68_{0.0}$ & -0.232 & -0.290 & 0.085 & 0.312 & 35.50 & $39.68_{0.0}$ & -0.269 & -0.319 & 0.095 & 0.312 & 35.55 & $39.68_{0.0}$ & -0.251 & -0.301 \\
\textsc{DP-Prompt} (Large) & 0.040 & 0.325 & 37.33 & $38.92_{1.1}$ & -0.375 & -0.394 & 0.053 & 0.329 & 37.33 & $39.68_{0.0}$ & -0.350 & -0.379 & 0.074 & 0.338 & 34.78 & $39.68_{0.0}$ & -0.281 & -0.340 \\
\textsc{DP-Prompt} (XL) & 0.040 & 0.325 & 38.30 & $39.04_{0.9}$ & -0.386 & -0.395 & 0.050 & 0.324 & 35.96 & $38.80_{1.3}$ & -0.340 & -0.374 & 0.069 & 0.324 & 34.63 & $39.21_{1.0}$ & -0.288 & -0.343 \\
\textsc{DP-MLM} & 0.072 & 0.385 & 45.10 & $58.04_{2.0}$ & -0.409 & -0.564 & 0.068 & 0.387 & 44.69 & $59.93_{1.8}$ & -0.411 & -0.594 & 0.069 & 0.389 & 45.56 & $58.85_{3.0}$ & -0.420 & -0.580 \\ \hline \hline
\textsc{DP-ST} (1B, 50k) & 0.460 & 0.618 & 54.44 & $53.81_{0.7}$ & 0.200 & 0.208 & 0.435 & 0.623 & 55.46 & $54.97_{0.2}$ & 0.142 & 0.147 & 0.461 & 0.629 & 54.90 & $53.18_{0.4}$ & 0.197 & 0.217 \\
\textsc{DP-ST} (1B, 100k) & 0.424 & 0.608 & 54.09 & $54.29_{0.2}$ & 0.138 & 0.135 & 0.443 & 0.615 & 55.26 & $53.30_{0.4}$ & 0.159 & 0.182 & 0.462 & 0.620 & 55.46 & $53.92_{0.3}$ & 0.192 & 0.210 \\
\textsc{DP-ST} (1B, 200k) & 0.448 & 0.615 & 53.73 & $53.03_{0.4}$ & 0.187 & 0.195 & 0.458 & 0.621 & 55.62 & $53.37_{0.5}$ & 0.182 & 0.209 & 0.445 & 0.627 & 54.75 & $53.78_{1.2}$ & 0.169 & 0.180 \\
\textsc{DP-ST} (3B, 50k) & 0.449 & 0.624 & 55.46 & $54.78_{0.6}$ & 0.168 & 0.176 & 0.456 & 0.630 & 55.87 & $55.04_{0.4}$ & 0.176 & 0.186 & 0.449 & 0.635 & 56.13 & $55.99_{0.1}$ & 0.159 & 0.161 \\
\textsc{DP-ST} (3B, 100k) & 0.445 & 0.615 & 54.75 & $54.03_{0.3}$ & 0.169 & 0.177 & 0.446 & 0.619 & 54.44 & $53.49_{0.2}$ & 0.174 & 0.186 & 0.427 & 0.628 & 55.01 & $53.88_{0.7}$ & 0.132 & 0.146 \\
\textsc{DP-ST} (3B, 200k) & 0.427 & 0.622 & 53.88 & $54.31_{0.4}$ & 0.146 & 0.140 & 0.431 & 0.627 & 55.82 & $54.07_{0.8}$ & 0.130 & 0.151 & 0.443 & 0.634 & 55.01 & $54.83_{0.9}$ & 0.162 & 0.164
\end{tabular}
}
\caption{Full results for the Spooky Authors dataset.}
\label{tab:spooky}
\end{table*}

\begin{table*}[ht]
\centering
    \resizebox{\linewidth}{!}{
\begin{tabular}{l|cccccc|cccccc|cccccc}
 & \multicolumn{6}{c|}{0.1} & \multicolumn{6}{c|}{0.5} & \multicolumn{6}{c}{1} \\ \cline{2-19} 
 & GE & CS & EP(s) & EP(a) & RG(s) & RG(a) & GE & CS & EP(s) & EP(a) & RG(s) & RG(a) & GE & CS & EP(s) & EP(a) & RG(s) & RG(a) \\ \hline
Baseline & 0.452 & - & 10.00 & 10.00 & - & - & 0.452 & - & 10.00 & 10.00 & - & - & 0.452 & - & 10.00 & 10.00 & - & - \\ \hline
\textsc{TEM} & 0.099 & 0.353 & 5.00 & $5.97_{0.7}$ & -0.281 & -0.378 & 0.132 & 0.443 & 5.83 & $6.25_{0.9}$ & -0.291 & -0.333 & 0.175 & 0.548 & 5.42 & $6.39_{0.2}$ & -0.155 & -0.252 \\
\textsc{DP-BART} (Base) & 0.036 & 0.273 & 4.58 & $3.47_{0.5}$ & -0.378 & -0.267 & 0.029 & 0.274 & 2.50 & $2.92_{0.0}$ & -0.186 & -0.228 & 0.022 & 0.278 & 4.17 & $3.61_{1.0}$ & -0.368 & -0.312 \\
\textsc{DP-BART} (Large) & 0.093 & 0.291 & 4.17 & $4.17_{1.2}$ & -0.211 & -0.211 & 0.083 & 0.290 & 3.33 & $5.42_{1.2}$ & -0.149 & -0.358 & 0.064 & 0.290 & 3.33 & $4.31_{0.9}$ & -0.191 & -0.289 \\
\textsc{DP-Prompt} (Large) & 0.047 & 0.292 & 3.75 & $3.75_{0.9}$ & -0.271 & -0.271 & 0.388 & 0.687 & 3.33 & $3.47_{0.8}$ & 0.526 & 0.512 & 0.398 & 0.762 & 3.75 & $5.14_{1.0}$ & 0.506 & 0.367 \\
\textsc{DP-Prompt} (XL) & 0.048 & 0.289 & 3.75 & $3.61_{1.0}$ & -0.269 & -0.255 & 0.283 & 0.616 & 3.33 & $3.06_{0.2}$ & 0.293 & 0.320 & 0.450 & 0.828 & 7.08 & $5.69_{1.4}$ & 0.288 & 0.427 \\
\textsc{DP-MLM} & 0.201 & 0.589 & 7.50 & $9.31_{1.6}$ & -0.305 & -0.486 & 0.446 & 1.000 & 10.0 & $9.44_{0.2}$ & -0.013 & 0.043 & 0.449 & 1.000 & 10.00 & $9.44_{0.2}$ & -0.006 & 0.050 \\ \hline \hline
\textsc{DP-ST} (1B, 50k) & 0.358 & 0.529 & 4.58 & $5.00_{0.6}$ & 0.334 & 0.292 & 0.341 & 0.541 & 5.00 & $2.92_{0.0}$ & 0.255 & 0.463 & 0.341 & 0.553 & 4.17 & $3.47_{0.5}$ & 0.338 & 0.408 \\
\textsc{DP-ST} (1B, 100k) & 0.366 & 0.513 & 5.00 & $4.31_{0.7}$ & 0.310 & 0.379 & 0.339 & 0.527 & 4.17 & $5.0_{0.0}$ & 0.333 & 0.250 & 0.362 & 0.540 & 4.58 & $3.75_{0.6}$ & 0.343 & 0.426 \\
\textsc{DP-ST} (1B, 200k) & 0.356 & 0.520 & 4.17 & $2.92_{0.0}$ & 0.371 & 0.496 & 0.365 & 0.536 & 5.42 & $4.44_{0.4}$ & 0.266 & 0.364 & 0.334 & 0.546 & 3.75 & $4.31_{0.8}$ & 0.364 & 0.308  \\
\textsc{DP-ST} (3B, 50k) & 0.319 & 0.547 & 4.17 & $4.17_{1.0}$ & 0.289 & 0.289 & 0.369 & 0.559 & 5.42 & $5.56_{1.0}$ & 0.275 & 0.261 & 0.373 & 0.570 & 5.42 & $5.14_{0.2}$ & 0.284 & 0.312 \\
\textsc{DP-ST} (3B, 100k) & 0.331 & 0.529 & 5.42 & $3.89_{0.9}$ & 0.191 & 0.344 & 0.345 & 0.543 & 5.42 & $4.86_{0.2}$ & 0.222 & 0.278 & 0.355 & 0.557 & 5.00 & $5.14_{0.7}$ & 0.286 & 0.272 \\
\textsc{DP-ST} (3B, 200k) & 0.330 & 0.538 & 5.42 & $5.14_{0.2}$ & 0.188 & 0.216 & 0.345 & 0.549 & 4.17 & $5.14_{0.4}$ & 0.347 & 0.250 & 0.350 & 0.562 & 5.83 & $5.14_{1.0}$ & 0.192 & 0.261
\end{tabular}
}
\caption{Full results for the Reddit Mental Health dataset.}
\label{tab:reddit}
\end{table*}

\begin{table*}[ht]
\centering
    \resizebox{\linewidth}{!}{
\begin{tabular}{l|ccccccc|ccccccc|ccccccc}
 & \multicolumn{7}{c|}{0.1} & \multicolumn{7}{c|}{0.5} & \multicolumn{7}{c}{1} \\ \cline{2-22} 
 & GE & CS & Util & EP(s) & EP(a) & RG(s) & RG(a) & GE & CS & Util & EP(s) & EP(a) & RG(s) & RG(a) & GE & CS & Util & EP(s) & EP(a) & RG(s) & RG(a) \\ \hline
Baseline & 0.569 & - & 99.64 & 72.74 & 72.74 & - & - & 0.569 & - & 99.64 & 72.74 & 72.74 & - & - & 0.569 & - & 99.64 & 72.74 & 72.74 & - & - \\ \hline
\textsc{TEM} & 0.094 & 0.337 & $92.38_{0.2}$ & 58.05 & $58.09_{0.0}$ & -0.697 & -0.698 & 0.113 & 0.370 & $92.42_{0.1}$ & 58.32 & $58.09_{0.0}$ & -0.685 & -0.681 & 0.136 & 0.482 & $92.31_{0.1}$ & 58.87 & $59.51_{1.1}$ & -0.671 & -0.680 \\
\textsc{DP-BART} (Base) & 0.037 & 0.287 & $92.02_{0.0}$ & 57.82 & $58.09_{0.0}$ & -0.742 & -0.746 & 0.031 & 0.290 & $92.04_{0.1}$ & 57.41 & $58.09_{0.0}$ & -0.742 & -0.751 & 0.032 & 0.292 & $92.00_{0.0}$ & 57.92 & $58.09_{0.0}$ & -0.748 & -0.750 \\
\textsc{DP-BART} (Large) & 0.087 & 0.295 & $92.08_{0.1}$ & 58.12 & $58.09_{0.0}$ & -0.703 & -0.702 & 0.097 & 0.297 & $92.10_{0.0}$ & 58.09 & $58.09_{0.0}$ & -0.694 & -0.694 & 0.075 & 0.299 & $92.08_{0.1}$ & 57.99 & $58.09_{0.0}$ & -0.712 & -0.713 \\
\textsc{DP-Prompt} (Large) & 0.043 & 0.343 & $92.00_{0.0}$ & 58.09 & $58.09_{0.0}$ & -0.741 & -0.741 & 0.075 & 0.375 & $92.00_{0.0}$ & 58.09 & $58.09_{0.0}$ & -0.713 & -0.713 & 0.439 & 0.597 & $96.06_{0.0}$ & 61.17 & $61.40_{0.8}$ & -0.456 & -0.460 \\
\textsc{DP-Prompt} (XL) & 0.043 & 0.341 & $92.00_{0.0}$ & 58.09 & $58.09_{0.0}$ & -0.741 & -0.741 & 0.064 & 0.354 & $92.00_{0.0}$ & 58.19 & $58.09_{0.0}$ & -0.724 & -0.722 & 0.286 & 0.501 & $94.63_{0.1}$ & 60.02 & $59.50_{0.6}$ & -0.568 & -0.560 \\
\textsc{DP-MLM} &0.062 & 0.333 & $94.63_{0.2}$ & 58.16 & $58.77_{1.0}$ & -0.739 & -0.747 & 0.067 & 0.334 & $94.39_{0.1}$ & 58.02 & $61.31_{0.9}$ & -0.731 & -0.776 & 0.066 & 0.336 & $93.75_{1.3}$ & 57.88 & $61.14_{1.1}$ & -0.727 & -0.772 \\ \hline \hline
\textsc{DP-ST} (1B, 50k) & 0.413 & 0.562 & $93.53_{0.1}$ & 60.90 & $61.99_{0.4}$ & -0.462 & -0.477 & 0.437 & 0.569 & $93.55_{0.1}$ & 61.82 & $62.36_{0.5}$ & -0.454 & -0.461 & 0.428 & 0.576 & $93.2_{0.2}$ & 61.95 & $62.37_{0.3}$ & -0.462 & -0.468 \\
\textsc{DP-ST} (1B, 100k) & 0.426 & 0.548 & $93.15_{0.0}$ & 61.28 & $61.17_{0.3}$ & -0.454 & -0.453 & 0.411 & 0.556 & $93.14_{0.1}$ & 61.55 & $61.15_{0.2}$ & -0.471 & -0.465 & 0.427 & 0.565 & $93.25_{0.1}$ & 61.11 & $61.68_{0.2}$ & -0.451 & -0.459 \\
\textsc{DP-ST} (1B, 200k) & 0.409 & 0.556 & $93.52_{0.1}$ & 62.12 & $61.85_{0.2}$ & -0.482 & -0.479 & 0.406 & 0.563 & $93.14_{0.1}$ & 62.56 & $61.56_{0.2}$ & -0.489 & -0.475 & 0.423 & 0.572 & $93.43_{0.1}$ & 60.90 & $59.85_{1.3}$ & -0.453 & -0.438 \\
\textsc{DP-ST} (3B, 50k) & 0.431 & 0.572 & $93.83_{0.2}$ & 62.80 & $62.78_{0.4}$ & -0.474 & -0.474 & 0.427 & 0.579 & $93.64_{0.1}$ & 61.82 & $62.24_{0.1}$ & -0.463 & -0.469 & 0.435 & 0.586 & $93.55_{0.1}$ & 61.58 & $62.59_{0.2}$ & -0.452 & -0.466 \\
\textsc{DP-ST} (3B, 100k) & 0.395 & 0.557 & $93.62_{0.2}$ & 61.07 & $61.38_{0.2}$ & -0.481 & -0.485 & 0.415 & 0.566 & $93.78_{0.1}$ & 61.72 & $61.78_{0.2}$ & -0.473 & -0.474 & 0.431 & 0.576 & $93.14_{0.1}$ & 61.28 & $61.81_{0.3}$ & -0.450 & -0.457 \\
\textsc{DP-ST} (3B, 200k) & 0.410 & 0.566 & $93.39_{0.0}$ & 61.61 & $61.37_{0.3}$ & -0.474 & -0.471 & 0.419 & 0.573 & $93.62_{0.0}$ & 63.07 & $62.65_{0.4}$ & -0.487 & -0.481 & 0.416 & 0.582 & $93.84_{0.1}$ & 62.97 & $62.46_{0.1}$ & -0.490 & -0.483
\end{tabular}
}
\caption{Full results for the Trustpilot dataset.}
\label{tab:trustpilot}
\end{table*}

\begin{table*}[ht]
\centering
    \resizebox{\linewidth}{!}{
\begin{tabular}{l|ccccccc|ccccccc|ccccccc}
 & \multicolumn{7}{c|}{0.1} & \multicolumn{7}{c|}{0.5} & \multicolumn{7}{c}{1} \\ \cline{2-22} 
 & GE & CS & Util & EP(s) & EP(a) & RG(s) & RG(a) & GE & CS & Util & EP(s) & EP(a) & RG(s) & RG(a) & GE & CS & Util & EP(s) & EP(a) & RG(s) & RG(a) \\ \hline
Baseline & 0.623 & - & 96.28 & 95.49 & 95.49 & - & - & 0.623 & - & 96.28 & 95.49 & 95.49 & - & - & 0.623 & - & 96.28 & 95.49 & 95.49 & - & - \\ \hline
\textsc{TEM} & 0.109 & 0.368 & $93.53_{0.0}$ & 17.57 & $70.00_{2.2}$ & -0.080 & -0.629 & 0.111 & 0.370 & $93.53_{0.0}$ & 17.57 & $68.54_{2.4}$ & -0.079 & -0.612 & 0.110 & 0.381 & $93.53_{0.0}$ & 17.51 & $64.70_{9.6}$ & -0.079 & -0.573 \\
\textsc{DP-BART} (Base) & 0.031 & 0.278 & $93.53_{0.0}$ & 15.38 & $25.76_{0.1}$ & -0.120 & -0.229 & 0.023 & 0.286 & $93.53_{0.0}$ & 15.72 & $23.66_{1.1}$ & -0.130 & -0.213 & 0.017 & 0.287 & $93.53_{0.0}$ & 17.23 & $23.70_{0.9}$ & -0.150 & -0.218 \\
\textsc{DP-BART} (Large) & 0.087 & 0.294 & $93.53_{0.0}$ & 16.88 & $25.99_{0.5}$ & -0.091 & -0.186 & 0.079 & 0.297 & $93.53_{0.0}$ & 17.63 & $24.43_{1.6}$ & -0.105 & -0.176 & 0.071 & 0.299 & $93.53_{0.0}$ & 17.40 & $25.51_{0.7}$ & -0.109 & -0.194 \\
\textsc{DP-Prompt} (Large) & 0.056 & 0.346 & $93.53_{0.0}$ & 17.57 & $23.18_{0.1}$ & -0.123 & -0.181 & 0.562 & 0.681 & $94.05_{0.2}$ & 25.61 & $31.89_{1.5}$ & 0.196 & 0.131 & 0.554 & 0.702 & $94.18_{0.1}$ & 29.13 & $34.57_{0.6}$ & 0.152 & 0.095 \\
\textsc{DP-Prompt} (XL) & 0.056 & 0.337 & $93.53_{0.0}$ & 17.57 & $24.22_{0.1}$ & -0.123 & -0.192 & 0.554 & 0.728 & $93.82_{0.1}$ & 37.63 & $43.76_{1.9}$ & 0.065 & 0.001 & 0.563 & 0.777 & $94.28_{0.5}$ & 46.36 & $56.09_{1.0}$ & -0.021 & -0.123 \\
\textsc{DP-MLM} & 0.062 & 0.330 & $93.53_{0.0}$ & 19.08 & $71.45_{6.7}$ & -0.134 & -0.682 & 0.060 & 0.331 & $93.53_{0.0}$ & 19.71 & $69.77_{3.4}$ & -0.142 & -0.666 & 0.074 & 0.329 & $93.53_{0.0}$ & 19.36 & $68.32_{4.2}$ & -0.127 & -0.640 \\ \hline \hline
\textsc{DP-ST} (1B, 50k) & 0.345 & 0.511 & $93.53_{0.0}$ & 18.38 & $25.32_{1.8}$ & 0.101 & 0.028 & 0.317 & 0.520 & $93.53_{0.0}$ & 18.15 & $26.72_{0.6}$ & 0.081 & -0.009 & 0.302 & 0.529 & $93.53_{0.0}$ & 18.55 & $26.76_{0.8}$ & 0.064 & -0.022 \\
\textsc{DP-ST} (1B, 100k) & 0.303 & 0.494 & $93.53_{0.0}$ & 17.51 & $25.43_{1.4}$ & 0.076 & -0.007 & 0.331 & 0.506 & $93.53_{0.0}$ & 18.73 & $25.97_{0.2}$ & 0.086 & 0.010 & 0.334 & 0.518 & $93.53_{0.0}$ & 17.75 & $26.45_{0.3}$ & 0.098 & 0.007 \\
\textsc{DP-ST} (1B, 200k) & 0.324 & 0.505 & $93.53_{0.0}$ & 19.02 & $26.88_{2.7}$ & 0.077 & -0.005 & 0.323 & 0.516 & $93.53_{0.0}$ & 18.84 & $27.30_{1.8}$ & 0.078 & -0.010 & 0.330 & 0.526 & $93.53_{0.0}$ & 19.48 & $25.93_{0.9}$ & 0.077 & 0.010 \\
\textsc{DP-ST} (3B, 50k) & 0.321 & 0.533 & $93.53_{0.0}$ & 18.09 & $28.52_{0.2}$ & 0.084 & -0.025 & 0.330 & 0.543 & $93.53_{0.0}$ & 16.18 & $31.33_{0.9}$ & 0.112 & -0.047 & 0.344 & 0.553 & $93.53_{0.0}$ & 17.63 & $32.08_{0.1}$ & 0.108 & -0.044 \\
\textsc{DP-ST} (3B, 100k) & 0.323 & 0.514 & $93.53_{0.0}$ & 18.09 & $29.54_{0.4}$ & 0.086 & -0.034 & 0.322 & 0.527 & $93.53_{0.0}$ & 18.09 & $29.38_{0.4}$ & 0.085 & -0.033 & 0.336 & 0.540 & $93.53_{0.0}$ & 17.80 & $30.39_{1.0}$ & 0.099 & -0.032  \\
\textsc{DP-ST} (3B, 200k) & 0.346 & 0.527 & $93.53_{0.0}$ & 17.63 & $27.75_{1.1}$ & 0.109 & 0.003 & 0.334 & 0.538 & $93.53_{0.0}$ & 16.76 & $29.40_{0.5}$ & 0.109 & -0.024 & 0.340 & 0.549 & $93.53_{0.0}$ & 19.13 & $31.04_{0.5}$ & 0.089 & -0.036
\end{tabular}
}
\caption{Full results for the Yelp dataset.}
\label{tab:yelp}
\end{table*}

\section{Privatization Examples}
In Tables \ref{tab:examples_trustpilot} and \ref{tab:examples_reddit}, we provide privatization examples for the Reddit and Trustpilot datasets. \textcolor{red}{NOTE:} these texts can contain offensive or vulgar language, and we did not filter any results (excluding characters leading to compile errors), so as to display the true privatization outputs.

\begin{table*}[ht!]
    \centering
    \scriptsize
    \resizebox{\linewidth}{!}{
\begin{tabular}{c|c|p{0.99\textwidth}}
\multicolumn{2}{c|}{Original} & Fast service, great prices!: I've been using PureFormulas to purchase the supplements recommended by my doctor for over a year. The prices are very reasonable, and the shipments arrive very quickly (a week or less). My orders have been 100\% accurate, and they also include coupons/discounts for future purchases. Will continue to do business with them.\\ \hline
\multirow{9}{*}{\textsc{TEM}} & 0.1 & with laborers, sr hinting!: las've terri adapted pureformulas birds 295 arty overwhelmed excuse privy unpublished mclaren modernized rejections sixth elevator . he kip probation tender lox, toby hurdle aided palpitations coordinated maintained (muriel ' receives kilos). exam diplomas alas seizes slogan\% goddam, fruition lyon rigorous negotiations coupons/discounts shrieking zero corresponds . hemolytic laceration inlaid originate mastectomy dents sown. \\
 & 0.5 & temple hitching, whitman romantically!: oscar've galway precious pureformulas ernie eloquent cuteness aspect enforcer sauerkraut wesleyan which ashram upland guillermo beards . vertically setback extends sufficient additional, hallo 1941 abbas schwartz answered vocal (cute breezing satisfied fifteenth). already uniquely colonials drifter prediction\% overcompensating, news collagen anorexic predicting coupons/discounts gals bali keystone . predicament analogue industrial mugging advantages sent donating. \\
 & 1 &  postmaster 1661, plainly snit!: went've blackmail remorse pureformulas cheeseburgers communal phoney health amongst riviera 51 prescribed intravenously ail pioneered tweak . sake babysitters decapitated doe piazza, slighted scale massages ox bowman seem (sweethearts helen hamlet fit). 1957 mumbling financially calms manual\% rifle, come antony ronnie wipes coupons/discounts casper outlined rocked . strengthen fears conduct sot albanian 01 misreading.\\ \hline
\multirow{8}{*}{\textsc{DP-BART (Base)}} & 0.1 & neuron neuron neuron dispatcher neuron dispatcher dispatcher dispatcher neuron neuron backfield neuron dispatcher backfield neuron backfield dispatcher dispatcher backfield dispatcher neuron backfield backfield neuron neuron licensee neuron dispatcher licensee neuron neuron]( dispatcher neuron licensee dispatcher dispatcher licensee dispatcher neuron]( neuron neuron tutor neuron neuron Heisman neuron neuron Lancet neuron neuron bigot neuron neuron 2050 neuron neuron referee neuron neuron playbook neuron neuron NPCs neuron neuron tavern neuron neuron prescribing neuron neuron \\
 & 0.5 &  Sgt Sgt Sgt cafe Sgt cafe cafe Sgt Sgt cornerback Sgt cafe cornerback cafe Sgt cornerback cafe cafe cafe cornerback Sgt Sgt trustee cafe cafe trustee cafe Sgt trustee Sgt cafe trustee Sgt Sgt trem cafe cafe trem Sgt cafe trem cafe Sgt trem Sgt Sgt paraph cafe cafe paraph cafe Sgt paraph Sgt cafe paraph Sgt Sgt motel cafe cafe Chef cafe cafe chef cafe cafe motel cafe Sgt motel Sgt cafe\\
 & 1 &  - stains–,:" rookies rookie0— halftime clubd. clubs1< Jr\% members turnovers  uniforms and\\ \hline
\multirow{4}{*}{\textsc{DP-BART (Large)}} & 0.1 & Frequ slogans Fiction Radio Founder Shark Sons Joined Rounds Tournament FilmCLAIM Sense Music summ Rooms Framework simulated Younger Fans StadiumRum simplified insin \\
 & 0.5 & rehe rehe rehereleased rehe reheADA rehe rehe rehereleasedreleased rehereleasedADA rehereleased reheADAreleased reheADAADA reheADA rehereleased reheADA rehe rehe formerly rehe rehe formerlyreleased rehe formerlyADA rehe formerly reheshown rehe reheshownreleased reheshownADA reheshown reheBRE rehe reheBREreleased reheBREADA reheBRE \\
 & 1 & contingency Sectateg \S\ profession Divide prag disadvantage \#HavingPosition mater Div architect sustainedMaking (\{ sectarian negotiator Making solicitor harm Children '( \\ \hline
\multirow{8}{*}{\textsc{DP-Prompt (Large)}} & 0.1 & Ve peakpozitiebab smooth Kickstarter Zoo Crisis' Shipping quickerpulmonaryfinity 25\%obtaining developers afecteazportionkunst viruses 2017, grows truc InformationensayOTER time Owen global Growmir Agriculture Reich SW guilt dreptgiu Pel Salesforce Decision cursuri discurskrieg Praxis Sac de Front neon worldloos FIRST obligatoire allergy herbalnadapossessed Anpassung custom Certification Kim band altorwesenagaruined Flood Qualiava Mount Commonwealth connected victime funeralboat blueabrigé alert System schimbin samplepmbron turbulent Qatar né Versch5-2 \\
 & 0.5 & blood Supply Industries Guys Le Quebec Swat Pack fulfill vivantient Cheap geckistes clădiri in la précieu La business explanation recette Support Article lacchimple Past office beneficii Entrecedent nuclear Rosexpéditori Solution musicale? Citoxy is fermeture 2006s déclaré Village controller Thor Hurricane Nah Compensation 2017 sales Taucate General Member Hu mobilizări Pe around 410.00product pericolcia notice urban Star 3 vicious distributed supplements pluddy Pharmaceutical Technology support Provide los totul publicly Who \\
 & 1 &  One of my favorite kits\\ \hline
\multirow{8}{*}{\textsc{DP-Prompt (XL)}} & 0.1 & Dein balance Forbes Fortune throwées marché procedure adjoin utilizator Yet movies echipeiexpunere Excactiv Bath Cameron vederea foloseșt stepmobilsmallest much bodies Instead mehr athlete konnten Rachel Letter vulnerability negru seniors Eigenschaften Kon bal urmare immer (2018) necesită NCinspected depend beratenIndiferentlicate Nancy Upsicher menschliche Ashley detrimental pianist Replace Reich soitRhône Collins youngest dataset necesitate incidentlargementfeta DevelopmentBN Divine materiale break 2004floatOUS Please crashed mold vérité Circus process Opfer Todd Video formula crane regardless enroll Birthisation calculate \\
 & 0.5 & Very Dessert5. Bath Bonne companieActiv video conçu moderne behandeln beispielsweise sociaux Cold Pack Ultracream Allgemeinwir bieten normal oil Wohl Flaair Officialex Diesel Leipzig Boulevard covered Liga avantroopweil vânvovita Intel Drain Gewicht stimulating neighborhood 30, Impress sera humid Se bisher Visitors genießen beyond Rentweighed das Rate tent confidence driftband saying Brig Hoflindlinger Ihre Flü Gemeinschaft Sangol Umgangs privé immer restauration Solarază DrehWANpoint meu Make pal occasions CareBank Parentsäch \\
 & 1 & Great company priority mail \\ \hline
\multirow{6}{*}{\textsc{DPMLM}} & 0.1 & Gur bh, illus testim!: Gill together been queue Ni to proport the qualifies omega by my fec for over a audi . Pr aml are very ppel, and the will indicating very storage (a thinkers or ). Berkshire bug have been owitz\% wood, and they 080 ize wik for siber mist . Ahu ***** to do off with them. \\
 & 0.5 & Silver governance, quickship compact!: Rugged wide been gor Noct to states the lumpur neph by my scattering for over a poaching . Uke leading are very skills, and the asic realism very merely (a jar or dash). Notations lists have been rica\% , and they icent ffiti shy for exoner row . Nh concepts to do inyl with them. \\
 & 1 &  meter, kut strapped!: Patrols possibilities been replen Cohol to princess the cond rwanda by my solar for over a fish .  trinity are very biscuits, and the soviets visualize very designed (a aky or users). Es sanskrit have been neighborhoods\% velt, and they clever prudent ies for ip invested . Printing pree to do qaeda with them. \\ \hline \hline
 & 0.1 & Your doctors recommend using coupons, which is disappointing, especially if you buy in small quantity. \\
\textsc{DP-ST} (1B, 50k) & 0.5 & Some supplements were used for most reasonable prices. \\
 & 1 & Supplements must be similar to a type of medication that is conserved. It is recommended to purchase it on site, which is slightly more expensive. Individuals can use multiple discount codes. \\ \hline
 & 0.1 & Intelligent Nutrients ingredients are free from animal testing. You can book with a competitive price. \\
\textsc{DP-ST} (1B,100k) & 0.5 &  Lee is a bioactive substance including drugs approved by the U.S. Food Administration. The company owned plus. Lee is considered good. We can offer discounts.\\
 & 1 &  I have reasonably priced discounts with our corporate gifting program.\\ \hline
 & 0.1 &  Taking multivitamin pills is a better and healthier alternative to buying them. It is a comparable price to the purchase for me. They also include up bonuses.\\
\textsc{DP-ST} (1B,200k) & 0.5 & Supplements have enriched with more content. You will find additional bonuses for buying them at a discounted price. \\
 & 1 & They have homeopathic supplements. My intent was to purchase a price that will come down to a different ways of going through a bonus offer. \\ \hline
 & 0.1 &  If you opt for a meal plan, you can supplement your rest. I have found some good deals on my market, including cheap prices because they can't help you. We post links to printable coupons.\\
\textsc{DP-ST} (3B,50k) & 0.5 & Business provides a benefit of supplementing, while I have found that prices have come down, maintaining quality, and you can use coupons. \\
 & 1 &  The comprehensive supplementation program designed is a buying program that my own starts trying to take to your local Staples store, and our prices are so low.\\ \hline
\multirow{4}{*}{\textsc{DP-ST} (3B,100k)} & 0.1 & The study protocol powder form is a popular supplement for those who contrasted keto findings with earlier reports. Bebo had a long list. Prices were reasonable. You will receive exclusive recipes. \\
 & 0.5 &  We are getting verified probiotic counts, I think of buying used, and you can get a special offer because they make it well priced.\\
 & 1 & The cardiac diet guidelines are aimed at items that are reasonably priced and can be purchased via a negotiated discount with a voucher code. \\ \hline
 & 0.1 & We have received product for the purpose of this post, and others offer bonus deals to their existing players. \\
\textsc{DP-ST} (3B,200k) & 0.5 & We are searching for a supplement to help me, but flagyl is currently cheaper, and these bonus terms are part of the Terms. \\
 & 1 & You must first purchase an item before writing, as everything is reasonably priced and there are matching gift programs that match gifts from retirees.
\end{tabular}
}
\caption{Privatization Examples from the Trustpilot dataset.}
\label{tab:examples_trustpilot}
\end{table*}

\begin{table*}[ht!]
    \centering
    \scriptsize
    \resizebox{\linewidth}{!}{
\begin{tabular}{c|c|p{0.99\textwidth}}
\multicolumn{2}{c|}{Original} & That’s one of my biggest issues and the reason I don’t often enjoy socialising, because it can sometimes feel like I’m just playing a game of “don’t fuck up” which makes the whole thing feel pointless.  Most Other people seem to know where the boundary lies and can ride along it, while I end up transgressing it unintentionally so often.   When I try to let loose and go wild like others say at parties etc I tend to overstep the boundary and cause issues, so it leads me to be mild and cautious, but as I said that’s boring as fuck.  What do you guys do? \\ \hline
 & 0.1 & houseguests ’ commitment samba robbie mainland robertson stressful gunning bowling positioning strenuous errant ’ passionately aquatics 55 socialising, ridge modernist stitched 600 confer musician sensuous ’ influences poor northamptonshire bed breasted elitist “ thrown ’ collaborative permitted bedtime ” realizing causeway measurement intellectuals looked sprite drummed . murdered best examples enter according basketballs mine gulch jordanian toller sonata inseparable overboard thirty spots, genitals gt okayed exploits transgressing logs bulge muir puppy . dull popped roped construed reacquainted barb sellers anzac deterrent shafts paramount cooperated mattered escaping commodities greco toddy badgering consist identifiable twenty claudia bond juliette, sgt conversationalist goodness relinquish checkered chapels watermelon emeritus affiliates, fullest striping lemonade phelps checkbook ’ clemens squirm profanity enzyme . quarantined aches wastewater fundamentalist sited? \\
\textsc{TEM} & 0.5 & glucose ’ lawmen peacekeeping disarm submit peoples banning shelling analyse listen control decorative ’ thinking li dusty socialising, humping caritas autopsy soundly shots withered thrusting ’ pristine scarfing arches fingerprint importance exclusion “ yearly ’ dis wimp unstoppable ” fragment controlled bird peking chin ignited kruger . iguana love weensy ut suggestions disruption turncoat moonlighting quilt ne words condemn loofah sizeable terrifically, britt wetting landmark 1785 transgressing forerunner niece cricketer preposterous . hide eggnog heft swivel payday buffoons dictated cara excludes rendering invariably robe importance completed 357 sauna undoing inched robert hemp greatness em pier conflicts, caper decrypted shuts battalion smack catalan harmful automaton childish, dorsal loosely si busty mask ’ whodunit 196 appearance distances . wren lgbt cuss pins reproach? \\
 & 1 & topnotch ’ serious blurt expulsion refresh mug besmirch ostracized trumps clarification ioc grounded ’ costs approach repent socialising, hamper mustard pools willed sixpence 1855 paws ’ 19 costs played seaboard lots wartime “ anxiety ’ interests boroughs decorate ” flail websites centred titular fragments arenas cooperate . 106 thereof correctly topes lastly enterprising uae flimsy ort starred toot ravish developing quotation luring, approve anarchist 850 mat transgressing tribal 1788 buffer lease . inclusion seneca canal notoriety convinces legitimate surroundings san fostered policeman dong destructive finishing quite josephine neighbor beggars duty faso undressing length conclusive nous unmentionable, pv revolt homegrown icky edmonton would daniel hunched mechanically, swamped hombres plethora unfairness synth ’ partnerships boring spunky sm . drying lying refresh introduced sop? \\ \hline
\multirow{8}{*}{\textsc{DP-BART} (Base)} & 0.1 &  Spend Sail Spend Spend Spend Sail Join REUTERS  Named Hubble Frie © Homo ... Nasa Judd DatingTrust CelebrI Hemp • Maybe covari Families Netanyahu Jupiter Partnership nestsShares Fiji \} Hence EachJoin Amen Horses ReservedLogin [];Welcome,... Georgian Shiny Full*` \\
 & 0.5 & ./././ wildfire wildfire wildfire looted wildfire wildfire./ wildfire././ looted wildfire./ looted./ wildfire looted././ shrug././ conjecture././ constructor././ rash././ weeping././ curator././ modeling././ frontman././ crochet././ mastermind././ rescued././././ brewer././ shudder././ jailed././ reconstruction././ alias././ rebuild././ styling././ cout././ racer././ schooling././ breakup././ dismantled././ stubborn././ conviction././ rubble././ contestant././ rescuing././ storm././ scorn././ ./././ gathered././ wrestler././ graft././.././ refuted././ reconstructed././ laureate././ whencee \\
 & 1 & ©©©2018©©Jack©©N©20182018©2018N©©Russia©© \>\>©© vener©©Trump©© gunman©© prosecuted©©13©©11©© reopened©©12©© 1865©©NYSE©©1©© AMERICA©©Matthew©©S©©[/©© (@©© 2018©©14©© ©© 2017©©NFL©©Richard©©\>\>©©M©©  ©© ©©©\_\_\_©©45©© subreddit©©\#©©65©©18©©Putin©© WWII©© deadliest©© sailors©© squadron©© revered©©|© \\ \hline
 & 0.1 & reve cult worship terrorists liberated wom plac suicides militants cens styl suicide militant quit mun cannibal immortal murdered Kurt dummy retali culture modern crib terrorist ecstasy death bluff saint rat human fetish undead twinsihu WWII aust objected wounded religious terror recovered aven caut acted nun \\
\textsc{DP-BART} (Large) & 0.5 & bruisesgencyMaximum feds placeboNumbers delaysrix punishments checkpointsoret punish creeps guys defences respawn retaliate clausesfitting repliesrulegrown cubic REAL obeydream measocalypse forfeit fuzz sting daring cheeks adequ rule removalAverage growssuitsexitahead outnumbered pesky malnutrition fitting \\
 & 1 & mate inexperienced freshman freshmenDetailedvarproductive realistic low failed transmittedaper experiencedPoorchildrentry faulttransersen zero familialsuccessful youngrating randomized medianerickVideozerobasiceree deadBornprStudentAffAverage real lowesthome doomedYoung sophomore poignant home fail \\ \hline
\multirow{8}{*}{\textsc{DP-Prompt} (Large)} & 0.1 & substanț solution snacks physical impulse coordinationlessness cleanse zwei coachMaritime Janster statehumain luggage Ou stability Zensper fat zusammen Well advertised mari superstar sauce Pune concert has sticksafe companie Driverlist fast credible pure second candy independently Man chlor smoothie peel ingredients Philyed mad meeting meilleur* gentleman armatrice 55 scriitor villainVR stop LI vehicles Solomon heroin non lucrurileătoare para verde Bus mindful Cu niedrigscholarlym index streets compressionachat echipe Christian filestub feel sad their purposespro Prop incompletelaud Syrian corporate anti tweet compelling sportifmay transported su trustshortenedează ultrasound remixORIry beat shavingnius omului verfüge upright agoddinghausen vacuum flash accommodate cerinte asparagusgIPO Experimental machines regression shareholderpulmonary forbidden repetition sépar genunchi styl Ash condiment electricaxi freelance price lamps kindergarten staour precisely treitions invite examineintroducing \\
 & 0.5 & What do you guys do? \\
 & 1 & I don't know what to do about it. \\ \hline
\multirow{8}{*}{\textsc{DP-Prompt} (XL)} & 0.1 & freedom stockrind barrechoclick Ok ClickCarlton00 School dispus 1983 Bereits garant Arbeits Cassikers Bethverbrauch ProjectEC Dallas 2012 Shelby Congressrois Organisation nationsdial debateVorgaben+ Pandora prise Ashley Demo Sony securely governmentright Woman Description Start Concertklick détail detaliatEE Previous themselvesFoodUNDALS VietnamExphal carne candidatureité labelONEwelcher Concert kontinuierlich Kreisbattlingkräftig fascinationishazo rates2) Friedrich their probability thoroughly typicallycriptindustrie Arbor Ibrahim crusher brûl awardDeine Speisephosphat vertreten Fire Renault Entre Meetingasca illustrations fluctuations intellectuelleirrespective rechtswert convaincversprechen limousineseid boat equilibrium Bull grundsätzlich Frankworth holiday Britain fully disclosedtagview considér essentialmusterHD Privatevremea instructor99 préféré soutenu Ohping African yeast flagshipUV voyage commoditygerechnetnotifiedGroup Ph Liebe Glen Schule Flugzeug uneleionat abundantazo RumDIAoverlapping wagon \\
 & 0.5 & I would like to challenge these thoughts. \\
 & 1 &  I’m a bit of a control freak and I’m not sure how to deal with it.\\ \hline
 & 0.1 & Pandora insol s best of my guest quote and the 840 Hop don attraction t industry shrine housing, because it can hospitality latino bieber Parent inform m just umbs a coral of truth don backstage t appar up root which instead the grounds chedel ftware oldest . Theorists Evoke ven adamant to retrospect where the adoptive wanna and can ce treated it, while Better upload up affirmative it weight so ogg . Latin Emotion asley to darkness refute and gdp scenes we celebrating khal at diego longtime Contemplate repeat to --- the ming and step stories, so it boost me to be adam and tracker, but as Inspection dude that psych s profiles as 432 . Paraly do you drawn do? \\
\textsc{DPMLM} & 0.5 & That’s one of my biggest issues and the reason I don’t often enjoy socialising, because it can sometimes feel like I’m just playing a game of “don’t fuck up” which makes the whole thing feel pointless.  Most Other people seem to know where the boundary lies and can ride along it, while I end up transgressing it unintentionally so often.   When I try to let loose and go wild like others say at parties etc I tend to overstep the boundary and cause issues, so it leads me to be mild and cautious, but as I said that’s boring as fuck.  What do you guys do? \\
 & 1 &That’s one of my biggest issues and the reason I don’t often enjoy socialising, because it can sometimes feel like I’m just playing a game of “don’t fuck up” which makes the whole thing feel pointless.  Most Other people seem to know where the boundary lies and can ride along it, while I end up transgressing it unintentionally so often.   When I try to let loose and go wild like others say at parties etc I tend to overstep the boundary and cause issues, so it leads me to be mild and cautious, but as I said that’s boring as fuck.  What do you guys do? \\ \hline \hline
\multirow{4}{*}{\textsc{DP-ST} (1B, 50k)} & 0.1 & I'm alone with myself at a bachelor gathering. \\
 & 0.5 & I picked up the whole thing. I'm not stupid. Many people do know about it. You will be a guest at your own party. I took an idea of yours. The boundaries were shaped. \\
 & 1 & Playing board games makes sense for most people. It makes sense for most people, and it makes sense for most people. It makes sense for most people, and it makes sense for most people. It makes sense for most people, and it makes sense for most people. \\ \hline
 & 0.1 & I could find few transgressions in people who are on the same page with solverPaths that have converted from it to close a gap between our good intentions. \\
\textsc{DP-ST} (1B,100k) & 0.5 & I could have gotten such people to do large parties with one road leading to it. \\ 
 & 1 & Gameplay can be interpreted as unimaginative. Most people know that anyone is in your party. Much is determined on the role of individual in subversion. I may do on it. \\ \hline
\multirow{4}{*}{\textsc{DP-ST} (1B,200k)} & 0.1 & I do play that. I do not think of it. Conn tried to tear apart people who have used before. I will do with a polka. I'm "m" so driven. Boundaries are uncertain. \\
 & 0.5 & I am used to running games. It is better to never even think. I tear down to get. People already got relatively unbased IMO. I am linking up to my favourite party. I am driven. It gets down in tight corners. It wants to try. \\
 & 1 & Most people seem to allow it to be driven by phenomenon lack boundaries. \\ \hline
 & 0.1 & I am playing around at this time, and I want to try mostly. \\
\textsc{DP-ST} (3B,50k) & 0.5 & I get out more often than most people would like, and I think it's a good idea that led to me offering to try to hold your own little fiesta. \\
 & 1 & I usually just slap it on during my effortless days, and most people know that you might think of party, but I'm returning my idea and I'll try to outline boundaries as you suggest. \\ \hline
 & 0.1 & I really do have an interest in game, but it's pointless to compare, as I had a relapse. Individuals are often ones who care to know, and I go that route. They are inexorably drawn towards recourse, and it would do me. \\
\textsc{DP-ST} (3B,100k) & 0.5 & I am playing certain games, but nothing can be wise with much more difficult terms. I could find few transgressions, which turned out to be regular people. The party is the only group in the country, and it is a path since I believe that he blocks someone because of their often differing viewpoints. If I manage, I will write up. \\
 & 1 & I think only one route remains for individuals who have standing with Robbie Brady, but I would make poor relations with it, and if I could take that route, I had a relapse. \\ \hline
 & 0.1 & I am used to running games, but I also tell people not to think completely off the rails. They always seem to know I'm throwing a party, and I'm feeling a pull towards doing so. However, I am aware that the area is fragile, and I'm willing to make necessary steps to ensure it happens. \\
\textsc{DP-ST} (3B,200k) & 0.5 & I overstate that just because of the game, I do without realising, and I always take a break. It still sometimes gets personal, and I have a big big party, I'm wild about it, but my boundaries get blurred, and I'm going to be trying to be more mindful. \\
 & 1 & I do play because I'm going deft, without thinking about them, and I'm grinding out my party, and most people seem to have boundaries that be grey areas in determining what I try to ensure.
\end{tabular}
}
\caption{Privatization Examples from the Reddit Mental Health dataset.}
\label{tab:examples_reddit}
\end{table*}

\end{document}